\documentclass{article}
\usepackage{ecj,epsfig,epstopdf,latexsym}
\usepackage{graphicx}
\usepackage{hyperref}
\usepackage{subfigure}

\DeclareMathSizes{12}{20}{14}{10}
\pdfoutput=1

\begin{document}

\title{The evolutionary origins of hierarchy } % For titles, only capitalize the first letter

\begin{center}
\author{Henok Mengistu, University of Wyoming \\
Joost Huizinga, University of Wyoming\\ 
Jean-Baptiste Mouret, Inria, Villers-l\`es-Nancy, F-54600, France\\ 
Jeff Clune, University of Wyoming}
\end{center}
\maketitle

%----------------------------------------------------------------------------------------

\begin{abstract}
Hierarchical organization -- the recursive composition of sub-modules -- is ubiquitous in biological networks, including neural, metabolic, ecological, and genetic regulatory networks, and in human-made systems, such as large organizations and the Internet. To date, most research on hierarchy in networks has been limited to quantifying this property. However, an open, important question in evolutionary biology is why hierarchical organization evolves in the first place. It has recently been shown that modularity evolves because of the presence of a cost for network connections. Here we investigate whether such connection costs also tend to cause a hierarchical organization of such modules. In computational simulations, we find that networks without a connection cost do not evolve to be hierarchical, even when the task has a hierarchical structure. However, with a connection cost, networks evolve to be both modular and hierarchical, and these networks exhibit higher overall performance and evolvability (i.e. faster adaptation to new environments). Additional analyses confirm that hierarchy independently improves adaptability after controlling for modularity. Overall, our results suggest that the same force--the cost of connections--promotes the evolution of both hierarchy and modularity, and that these properties are important drivers of network performance and adaptability. In addition to shedding light on the emergence of hierarchy across the many domains in which it appears, these findings will also accelerate future research into evolving more complex, intelligent computational brains in the fields of artificial intelligence and robotics. 
\end{abstract}

\section*{Introduction}

Hierarchy is an important organizational property in many biological and man-made systems, ranging from neural \cite{hierahicalmodularity,hierahicalmodularity2}, ecological \cite{ecosystem}, metabolic \cite{metabolic}, and genetic regulatory networks \cite{generegulatory}, to the organization of companies \cite{companyhierarchy}, cities \cite{urbanherarcy}, societies \cite{socialhierarchy}, and the Internet \cite{internethierarchy,ravasz2003hierarchical}.
There are many types of hierarchy \cite{mones2012:hierarchy,pumain2006hierarchy,hierarchy:polysemy}, but the one most relevant for biological networks~\cite{sales2007extracting}, especially neural networks \cite{hierahicalmodularity,hierahicalmodularity2,hierarchyembedding}, refers to a recursive organization of modules \cite{ravasz2003hierarchical,hierarchy:polysemy}. Modules are defined as highly connected clusters of entities that are only sparsely connected to entities in other clusters \cite{clune2013:evolutionary,lipson:principles,wagner2007:road}. Such hierarchy has long been recognized as a ubiquitous and beneficial design principle of both natural and man-made systems \cite{sales2007extracting}. For example, in complex biological systems, the hierarchical composition of modules is thought to confer greater robustness and adaptability~\cite{hierahicalmodularity,hierahicalmodularity2,brainhierarchy,hierarchyembedding}, whereas in engineered designs, a hierarchical organization of simple structures accelerates the design, production, and redesign of artifacts \cite{lipson:principles,hierarchy:engineering,hierarchy:circuits}.

While most studies of hierarchy focus on producing methods to quantify it \cite{mones2012:hierarchy,metabolic,internethierarchy,trusina2004hierarchy,corominas2011measuring,dehmer2008entropy,song2006origins,ryazanov1988dynamics,sales2007extracting}, a few have instead examined why hierarchy emerges in various systems. In some domains, the emergence of hierarchy is well understood; e.g., in complex systems, such as social networks, ecosystems, and road networks, the emergence of hierarchy can be explained solely by local decisions and or interactions ~\cite{corominas2013:origins,o1986hierarchical,wu2002spatially,ecosystem}. But, in biological systems, where the evolution of hierarchy is shaped by natural selection, why hierarchy evolves, and whether its evolution is due to direct or indirect selection, is an open and interesting question~\cite{ecosystem,open:question}.
Non-adaptive theories state that the hierarchy in some, but not all, types of biological networks may emerge as a by-product of random processes \cite{corominas2013:origins}.
Most adaptive explanations claim that hierarchy is directly selected for because it confers evolvability \cite{salthe2013evolving}, which is the ability of populations to quickly adapt to novel environments \cite{evovability:2008}. Yet in computational experiments that simulate natural evolution, hierarchy rarely, if ever, evolves on its own~\cite{clune2010investigating,paine2005hierarchical,huizingaevolving}, suggesting that alternate explanations are required to explain the evolutionary origins of hierarchy. Moreover, even if hierarchy, once present, is directly selected for because of the evolvability it confers, explanations are still required for how that hierarchy emerges in the first place. 

In this paper we investigate one such hypothesis: the existence of costs for network connections creates indirect selection for the evolution of hierarchy. This hypothesis is based on two lines of reasoning. The first is that hierarchy requires a recursive composition of modules \cite{ravasz2003hierarchical}, and the second is that hierarchy includes sparsity. A recent study demonstrated that both modularity and sparsity evolve because of the presence of a cost for network connections~\cite{clune2013:evolutionary}. Connection costs may therefore promote both modularity and sparsity, and thus may also promote the evolution of hierarchy.

It is realistic to incorporate connection costs into biological network models because it is known that there are costs to create connections, maintain them, and transmit information along them \cite{cherniak2004global,chen2006wiring}. 
Additionally, evidence supports the existence of a selection pressure in biological networks to minimize the net cost of connections. For example, multiple studies have shown that biological neural networks, which are hierarchical \cite{hierahicalmodularity,hierahicalmodularity2}, have been organized to reduce their amount of wiring by having fewer long connections and by locating neurons optimally to
reduce the wiring between them~\cite{chen2006wiring,raj2011wiring,ahn2006wiring,laughlin2003communication}.

A relationship between hierarchy and connection costs can also be observed in a variety of different man-made systems. For example, very large scale integrated circuits (VLSI), which are designed to minimize wiring, are hierarchically organized \cite{hierarchyembedding}. In organizations such as militaries and companies, a hierarchical communication model has been shown to be an ideal configuration when there is a cost for communication links between organization members \cite{R.Guimera:2001}. However, there is no prior work that tests whether the presence of connection costs is responsible for the evolution of hierarchy. Here we test that hypothesis in computational simulations of evolution and our experiments confirm that hierarchy does indeed evolve when there is a cost for network connections (Fig. 1).

\begin{table}[b]
\centering
\caption{The main problem (pictured in Fig.~\ref{fig:allresults}A). Networks receive 8-bit vectors as inputs. As shown, a successful network could AND adjacent input pairs, XOR the resulting pairs, and AND the result. Performance is a function only of the final output, and thus does not depend on how the network solves the problem; Other, non-hierarchical solutions also exist.}
\label{mainprotable}
\begin{tabular}{l l l}

\hline
\textbf{} & \textbf{Values} \\
\hline
Input pattern & 0 0 \hspace{0.01in}  1 1 \hspace{0.01in}  0 1 \hspace{0.01in}  1 1 \\
AND gate & \hspace{0.01in} 0 \hspace{0.16in}1\hspace{0.2in}0\hspace{0.2in}1 \\
XOR gate  &\hspace{0.16in} 1 \hspace{0.4in}1\\
AND gate  & \hspace{0.47in}1 \\
\hline
\end{tabular}
\end{table}

We also investigate the hypothesis that hierarchy confers evolvability, which has long been argued~\cite{hierahicalmodularity,hierahicalmodularity2,hierarchyembedding,simon:archicomplex}, but has not previously been extensively tested \cite{hierarchyembedding}. Our experiments confirm that hierarchical networks, evolved in response to connection costs, exhibit an enhanced ability to adapt.

\begin{figure}
\centering
\includegraphics[scale=0.25]{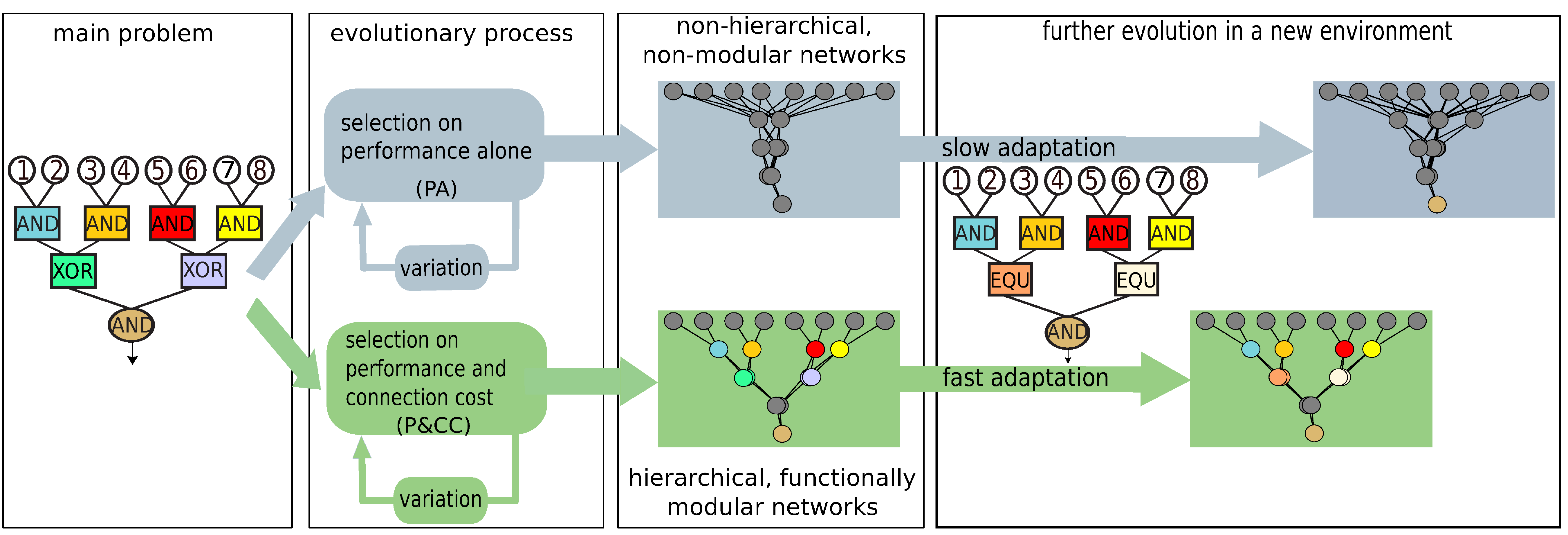}
\caption{The main hypothesis. Evolution with selection for performance only results in non-hierarchical and non-modular networks, which take longer to adapt to new environments. Evolving networks with a connection cost, however, creates hierarchical and functionally modular networks that can solve the overall problem by recursively solving its sub-problems. These networks also adapt to new environments faster.}
\label{fig:concept_figure}
\end{figure}
Experimentally investigating the evolution of hierarchy in biological networks is impractical, because natural evolution is slow and it is not currently possible to vary the cost of biological connections. Therefore, we conduct experiments in computational simulations of evolving networks. Computational simulations of evolution have shed substantial light on open, important questions in evolutionary biology~\cite{lenski1999genome,lenski2003evolutionary,wilke2001evolution}, including the evolution of modularity~\cite{clune2013:evolutionary,wagner2007:road,espinosa2010specialization,Kashtan2005,Kashtan2007}, a structural property closely related to hierarchy. In such simulations, randomly generated individuals recombine, mutate, and reproduce based on a fitness function that evaluates each individual according to how well they perform a task. The task can be analogous to efficiently metabolizing resources or performing a required behavior. This process of evolution cycles for a predetermined number of generations.

We evolved computational abstractions of animal brains called artificial neural networks (ANNs)  \cite{alon2007:introduction,trappenberg2010:fundamentals} to solve hierarchical Boolean logic problems (Fig.~\ref{fig:allresults}A). In addition to abstracting animal brains, ANNs have also been used as abstractions of gene regulatory networks \cite{geard2005gene}. They abstract both because they sense their environment through inputs and produce outputs, which can either be interpreted as regulating genes or moving muscles (Methods). In our experiments, we evolve the  ANNs with or without a cost for network connections. Specifically, the experimental treatment selects for maximizing performance and minimizing connection costs (\textit{performance and connection cost}, P\&CC), whereas the control treatment selects for performance only (\textit{performance alone}, PA).

In all treatments the evolving networks have eight inputs and a single output. During evaluation, each network is tested on all possible (256) input patterns of zeros and ones, and the network's output is checked against a hierarchical Boolean logic function provided with the same input (Fig. 2A and Table 1). An ANN output $\geq0$ is considered True and an output $<0$ is considered False. A network's performance (fitness) is its percent of correct answers over all input patterns.

\section*{Results}
\begin{figure}
\vspace{-175pt}\hspace{-10pt}
\includegraphics[scale=.73]{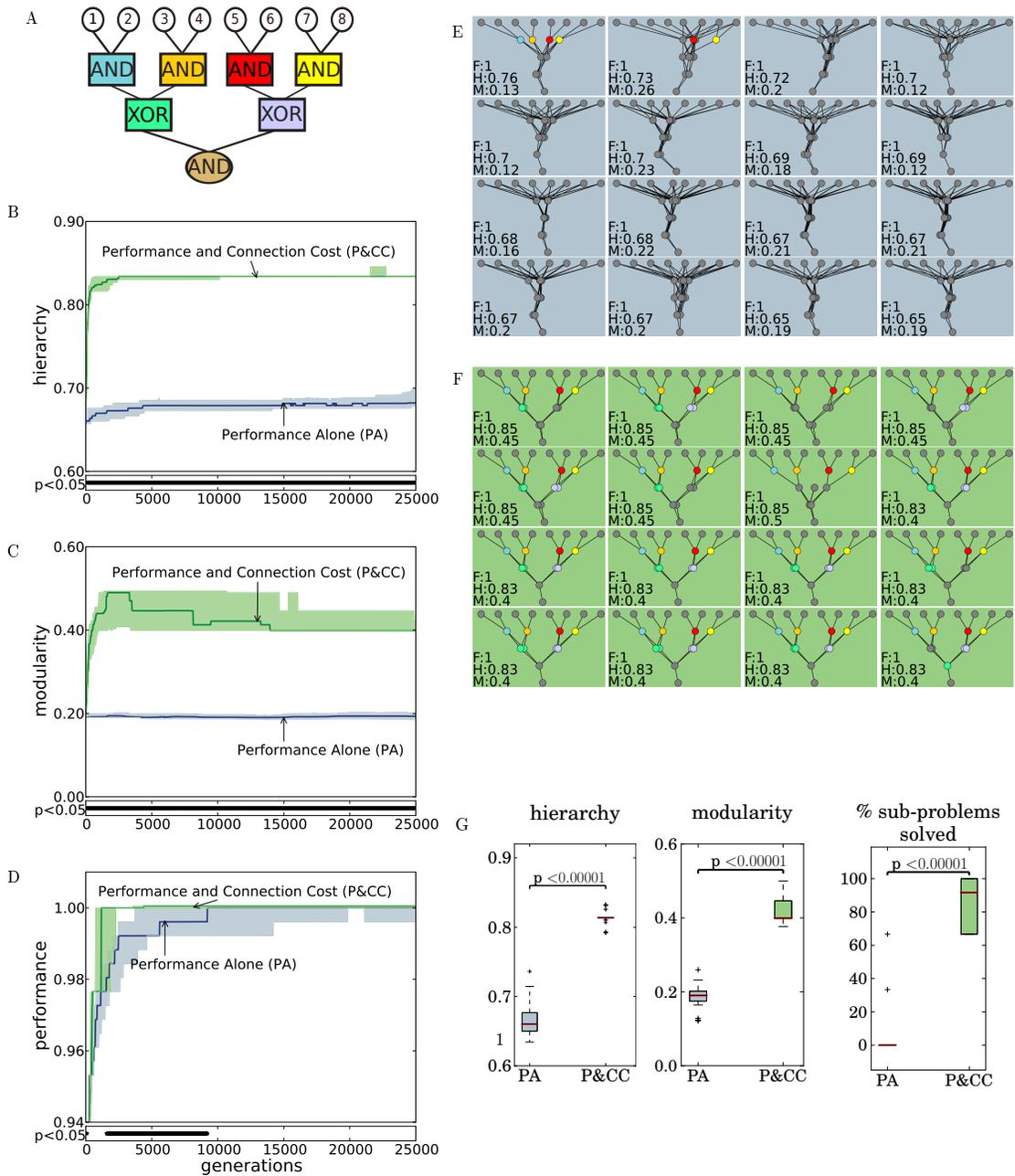}
\caption{A cost for network connections produces networks that are significantly more hierarchical, modular, high-performing, and likely to functionally decompose a problem. The algorithms for quantifying hierarchy and modularity are described in Methods. The bars below plots indicate at which generation a significant difference exists between the two treatments.~\textbf{(A)} The hierarchical AND-XOR-AND problem (the default for our experiments). The top eight nodes are inputs to the problem and the bottom node is an output. \textbf{(B)} P\&CC networks are significantly more hierarchical than PA networks. $p$-values are from the Mann-Whitney-Wilcoxon rank-sum test, which is the default statistical test throughout the paper unless otherwise stated. \textbf{(C)} P\&CC networks are also significantly more modular than PA networks, confirming a previous finding~\cite{clune2013:evolutionary,huizingaevolving}. \textbf{(D)} P\&CC networks evolve a solution to the problem significantly faster. \textbf{(E)} Evolved networks from the 16 highest-performing replicates in the PA treatment. The networks are non-hierarchical, non-modular, and do not tend to decompose the problem. Each network panel reports fitness/performance (F), hierarchy (H), and modularity (M). Nodes are colored if they solve one of the logic sub-functions in (\textbf{A}). Fig. S1 shows networks from all 30 replicates for both treatments. \textbf{(F)} Evolved networks from the 16 highest-performing replicates in the P\&CC treatment. The networks are hierarchical, modular, and decompose the problem. \textbf{(G)} A comparison of P\&CC and PA networks from the final generation. P\&CC networks are significantly more hierarchical, modular, and solve significantly more sub-problems.} 

\label{fig:allresults}

\end{figure}

On the main experimental problem (Fig.~\ref{fig:allresults}A), the addition of a connection cost leads to the evolution of significantly more hierarchical networks (Fig.~\ref{fig:allresults}B,G). Confirming previous findings on different problems \cite{clune2013:evolutionary,huizingaevolving}, the addition of a connection cost also significantly increases modularity (Fig.~\ref{fig:allresults}C,G) and reduces the number of generations required to evolve a solution (Fig.~\ref{fig:allresults}D). 
 
Importantly, while final performance levels for the performance and connection cost (P\&CC) treatment are similar to those of the performance alone (PA) treatment, there is a qualitative difference in how the networks solve the problem. P\&CC networks exhibit \emph{functional hierarchy} in that they solve the overall problem by recursively combining solutions to sub-problems (Fig.~\ref{fig:allresults}F), whereas the PA networks tend to combine all input information in the first layer and then process it in a monolithic fashion (Fig.~\ref{fig:allresults}E). Such functional hierarchy can be quantified as the percent of \emph{sub-problems} a network solves (e.g. the AND and XOR gates in Fig.~\ref{fig:allresults}A). A sub-problem is considered solved if, for all possible network inputs, and for any threshold, a neuron in a network exists that outputs an above-threshold value whenever the answer to the sub-problem is True, and a sub-threshold value when the answer is False, or vice-versa (Methods). This measure reveals that evolved P\&CC networks solve significantly more sub-problems than their PA counterparts (Fig.~\ref{fig:allresults}G, $p<2.6 \times 10^{-16}$ via Fisher's exact test).

\begin{figure*}[thb!]
\hspace{-13pt}
\includegraphics[scale=0.9]{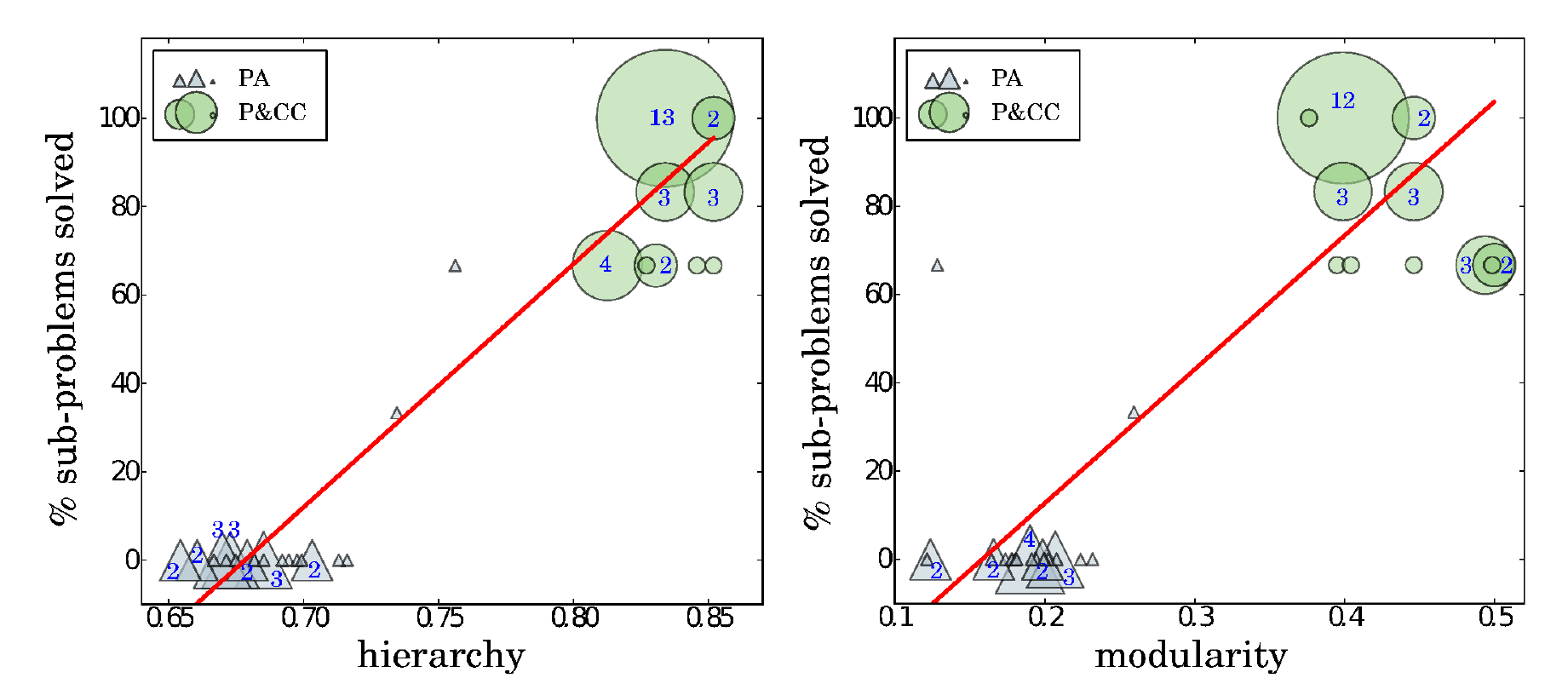}
\caption{Solving sub-problems is correlated with both hierarchy (left) and modularity (right). The shape sizes and enclosed numbers indicate the number of networks at that coordinate (an empty shape indicates only one network is present). The Pearson's correlation coefficient is 0.96 for hierarchy and 0.87 for modularity, indicating strong, linear, positive relationships. Both correlations are significant ($p < 0.00001$) according to a t-test with a correlation of zero as the null hypothesis.}
\label{fig:hierachyandsubproblemcorrelarion}
%\vspace{-3.0em}
\end{figure*}

To further investigate how the ability to solve sub-problems is related to hierarchy and modularity, we plotted the percent of sub-problems solved vs. both hierarchy and modularity (Fig.~\ref{fig:hierachyandsubproblemcorrelarion}). The plots show a significant, strong, positive correlation between the ability to solve sub-problems and both hierarchy and modularity. 

\begin{figure*}[thb]
\hspace{-13pt}
\centering
\includegraphics[scale=0.8]{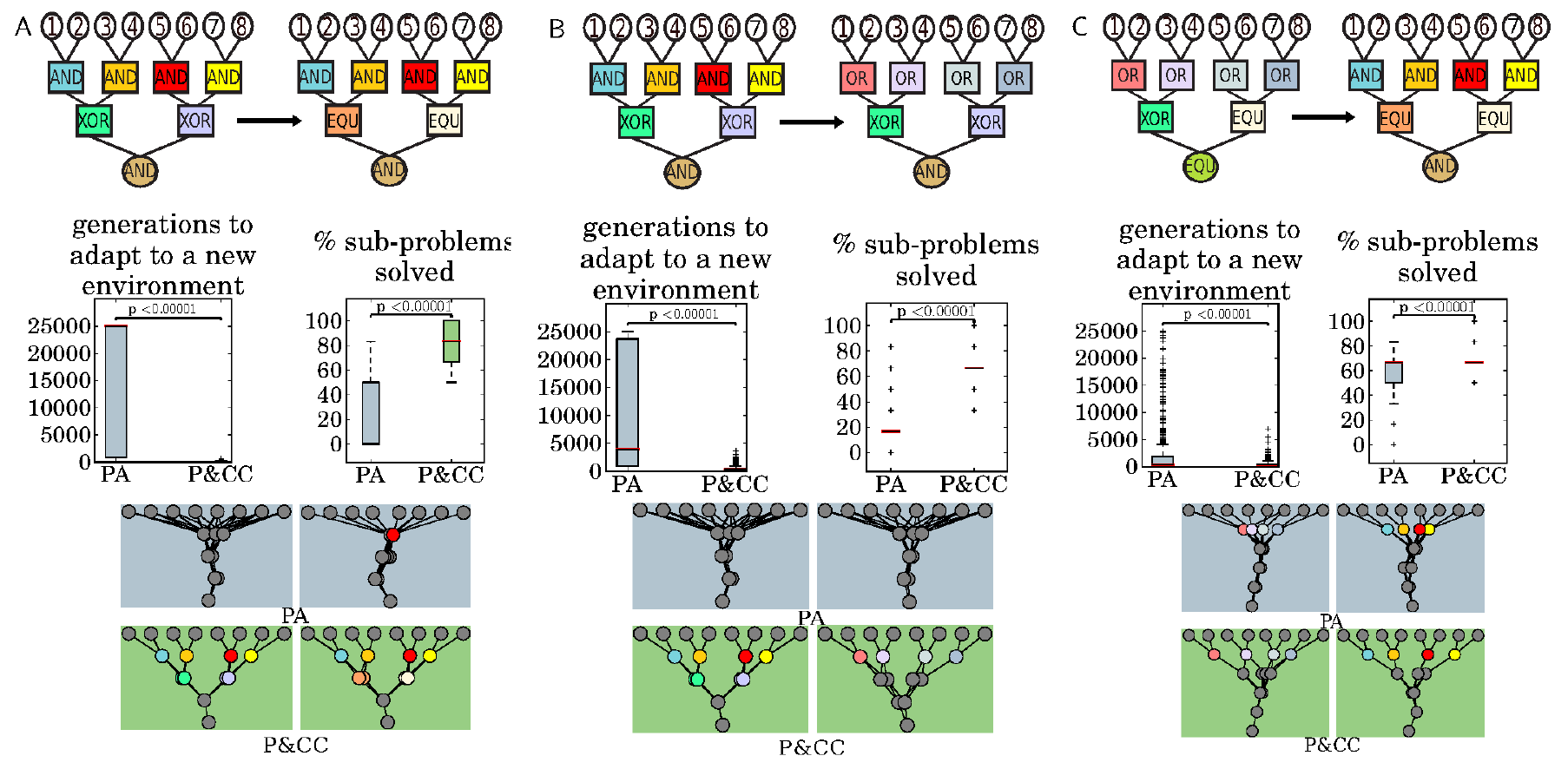}

\caption{P\&CC networks adapt significantly faster and solve significantly more sub-problems in new environments. In these experiments, networks first evolve to solve a problem perfectly in a base environment (left) and are then placed in a target environment (right) where they continue evolving to solve a different problem. The evolvability of PA and P\&CC networks is quantified as the number of generations they take to solve the new problem perfectly. A pair of evolved networks is shown for both treatments. The left one shows the network with median hierarchy (here and elsewhere, rounding up) of 30 replicates in the base environment; the right one shows the median hierarchy network of the 30 runs in the target environment started with the network on the left. Figs. S5-S10 show all network pairs.}
\label{fig:evolvability}
\end{figure*}

It has been hypothesized that one advantage of network hierarchy is that it confers evolvablity \cite{hierahicalmodularity,hierahicalmodularity2,hierarchyembedding}. We test this hypothesis by first evolving networks to solve one problem (the \emph{base environment}) and then evolving those networks to solve a different problem (the \emph{target environment}). To isolate evolvability, we keep initial performance equal by taking the first 30 runs of each treatment (PA and P\&CC) that evolve a perfectly-performing network for the base environment~\cite{clune2013:evolutionary}. Each of these 30 networks then seeds 30 runs in the target environment (for 900 total replicates per treatment). The base and target problems are both hierarchical and share some, but not all, sub-problems~(Fig.~\ref{fig:evolvability}). Evolution in the target environment continues until the new problem is solved or 25000 generations elapse. We quantify evolvability as the number of generations required to solve the target problem \cite{clune2013:evolutionary}. We performed three such experiments, each with different base and target problems. In all experiments, P\&CC networks take significantly fewer generations to adapt to the new environment than PA networks. They also solve significantly more of the target problem's sub-problems (Fig.~\ref{fig:evolvability}).

One possible reason for the fast adaptation of P\&CC networks is that their modular structure allows solutions to sub-problems to be re-used  in different contexts~\cite{clune2013:evolutionary}. A hierarchical structure may also be beneficial if both problems are hierarchical, even if the computation at points in the structure is different. For example, modules that solve XOR gates can quickly be rewired to solve EQU gates (Fig.~\ref{fig:evolvability}A). Another reason for faster P\&CC adaptation could be that these networks are sparser, meaning that fewer connections need to be optimized.

\begin{figure*}
\hspace{0pt}

\includegraphics[trim=2cm 17cm 0cm 6cm,scale=0.85]{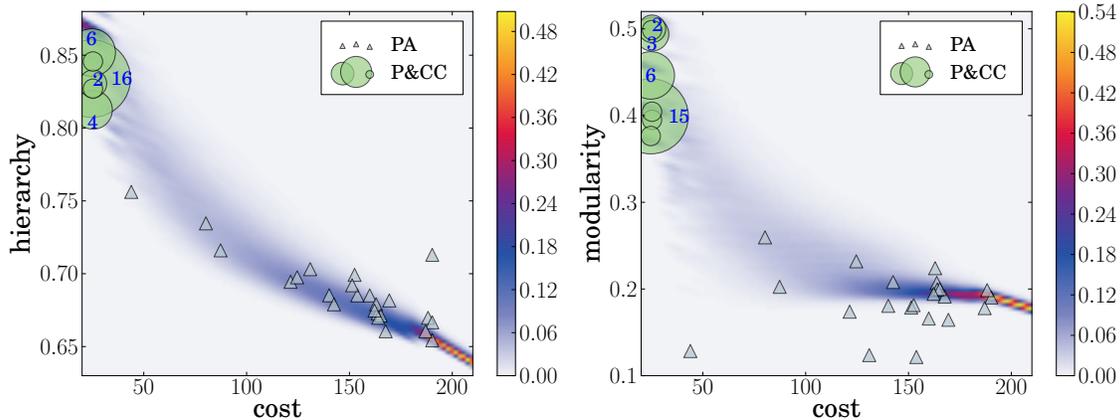}
\caption{ Lower cost networks are more hierarchical and modular. The hierarchy (left) and modularity (right) of randomly generated (i.e. non-functional) networks is shown for each cost after being normalized per cost value and then smoothed by a Gaussian kernel density estimation function. Colors indicate the probability of a network being generated at that location (heat map). Networks evolved in either the P\&CC or PA treatment are overlaid as green circles or blue triangles, respectively. Circle or triangle size and the enclosed number indicate the number of networks at that coordinate (no number means 1). All evolved P\&CC networks are in the high-hierarchy, low-cost region. Most evolved PA networks are in the high-cost, low-hierarchy region.}
 \label{fig:SamplingHierarchy}
 %\vspace{-0.4em}
\end{figure*}

To further understand why connection costs increase hierarchy, we generated 20000 random, valid networks for each number of connections a network could have (Methods). A network is valid if it has a path from each of the input nodes to the output node. The networks were neither evolved nor evaluated for performance. Of these networks, those that are low-cost tend to have high hierarchy, and those with a high cost have low hierarchy (Fig.~\ref{fig:SamplingHierarchy}, left). This inherent association between low connection costs and high hierarchy suggests why selecting for low-cost networks promotes the evolution of hierarchical networks. It also suggests why networks evolve to be non-hierarchical without a pressure to minimize connection costs. Indeed, most evolved PA networks reside in the high-cost, low-hierarchy region, whereas all P\&CC networks occupy the low-cost, high-hierarchy region (Fig.~\ref{fig:SamplingHierarchy} left). 

Similarly, there is also an inverse relationship between cost and modularity for these random networks (Fig.~\ref{fig:SamplingHierarchy}, right), as was shown in~\cite{clune2013:evolutionary}. All evolved P\&CC networks are found in the low-cost, high-modularity region; PA networks are spread over the low-modularity, high-cost region (Fig.~\ref{fig:SamplingHierarchy}, right).

\begin{figure*}
\vspace{-85pt} \hspace{-40pt}
\includegraphics[trim=0cm 12cm 0cm 6cm,scale=0.82]{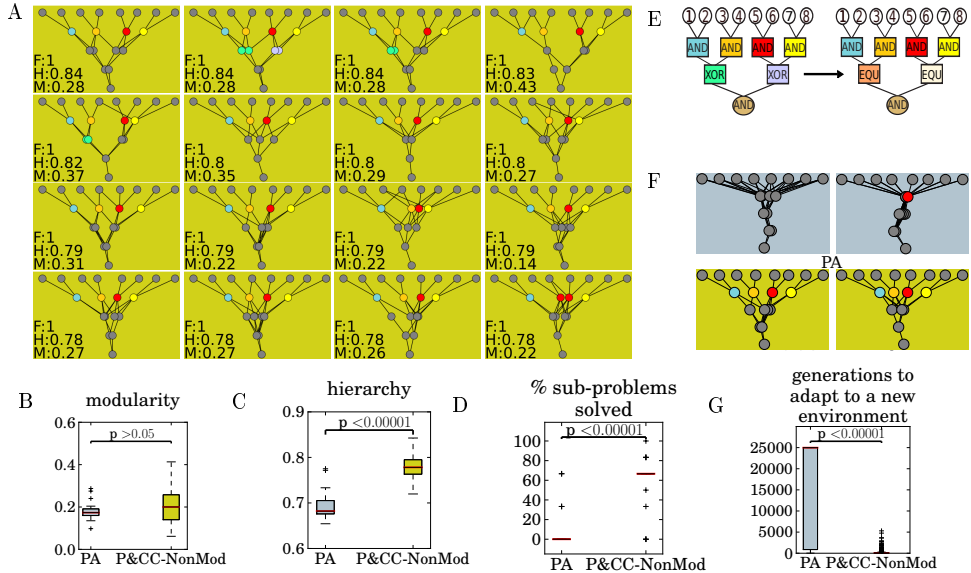}    
\caption{Evolving low-cost, high-performing networks that are non-modular reveals that independent of modularity, a connection cost promotes the evolution of hierarchy. \textbf{(A)} Networks from the 16 highest-performing P\&CC-NonMod replicates (Fig. S4 shows networks from all 30 trials). The networks are hierarchal, but not highly modular. \textbf{(B)} There is no significant difference in modularity between P\&CC-NonMod and PA networks, but P\&CC-NonMod networks are significantly more hierarchical~\textbf{(C)} and solve significantly more sub-problems~\textbf{(D)} than PA networks. \textbf{(E-G)} P\&CC-NonMod networks also adapt significantly faster to a new environment than PA networks, suggesting that hierarchy promotes evolvability independently of modularity. \textbf{(E)} The base and target problem for this evolvability experiment. \textbf{(F)} A perfect-performing network evolved for the base problem (left) and a descendant network evolved on the target problem (right). The example networks are those with median hierarchy: Fig. S11 shows all pairs.~\textbf{(G)} P\&CC-NonMod networks adapt significantly faster to the new problem.}
\label{fig:PCC-NonMod}
%\vspace{-6.em}
\end{figure*}

As Figs.~\ref{fig:hierachyandsubproblemcorrelarion} and \ref{fig:SamplingHierarchy} suggest, network modularity and hierarchy are highly correlated (Fig. S12, Pearson's correlation coefficient $=0.92$, $p < 0.00001$ based on a t-test with a correlation of zero as the null hypothesis). It is thus unclear whether hierarchy evolves as a direct consequence of connection costs, or if its evolution is a by-product of evolved modularity. To address this issue we ran an additional experiment where evolutionary fitness was a function of networks being \textit{low-cost} (as in the P\&CC treatment), \textit{high-performing} (as in all treatments), and \textit{non-modular} (achieved by selecting for low modularity scores). We call this treatment P\&CC-NonMod. The results reveal that P\&CC-NonMod networks have the same low level of modularity as PA networks (Fig.~\ref{fig:PCC-NonMod}A-B, $p = 0.23$),  but have significantly higher hierarchy (Fig.~\ref{fig:PCC-NonMod}A, C, $p<0.00001$) and solve significantly more sub-problems than PA networks~(Fig.~\ref{fig:PCC-NonMod}D). These results reveal that, independent of modularity, a connection cost promotes the evolution of hierarchy. Additionally, P\&CC-NonMod networks are significantly more evolvable than PA networks~(Fig.~\ref{fig:PCC-NonMod}E-G), revealing that hierarchy promotes evolvability independently of the known evolvability improvements caused by modularity~\cite{clune2013:evolutionary}. 

\begin{figure*}[htb!]
\centering
\includegraphics[scale=0.4]{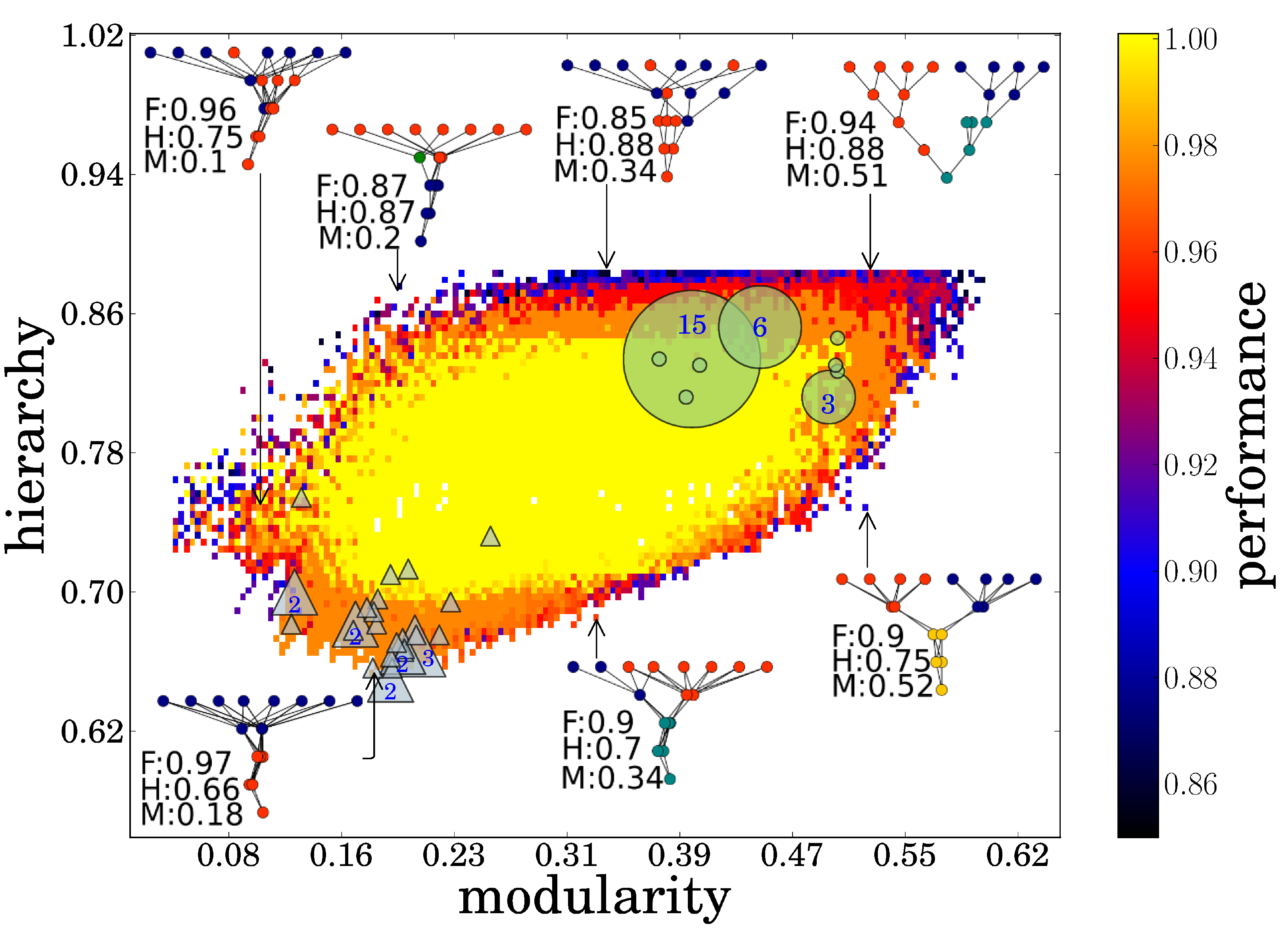}
\caption{ Network modularity and hierarchy can independently vary, and high-performing networks exist with a wide range of modularity and hierarchy scores.
The highest-performing networks evolution discovered (with the MAP-Elites algorithm) for each combination of modularity and hierarchy. A few example networks are shown, along with their fitness (F), hierarchy (H), and modularity (M). The best network from each of the PA and P\&CC treatments are also overlaid as blue triangles and green circles, respectively. The size of the circles or triangles and the enclosed number indicate the number of networks at that coordinate (no number means 1).}
\label{fig:map-elitesexp}
\end{figure*}

To gain better insight into the relationship between modularity, hierarchy, and performance, we searched for the highest-performing networks at all possible levels of modularity and hierarchy. We performed this search with the multi-dimensional archive of phenotypic elites (MAP-Elites) algorithm \cite{map-elite}. The results show that networks with the same modularity can have a wide range of different levels of hierarchy, and vice-versa, which indicates that these network properties can vary independently (Fig.~\ref{fig:map-elitesexp}). Additionally, the high-hierarchy, high-modularity region, in which evolved P\&CC networks reside, contains more high-performing solutions than the low-hierarchy, low-modularity region where PA networks reside (Fig.~\ref{fig:map-elitesexp}), suggesting an explanation for why P\&CC networks find high-performing solutions faster (Fig.~\ref{fig:allresults}D). 
\begin{figure*}[htb]
\hspace{-20pt}
\includegraphics[scale=0.96]{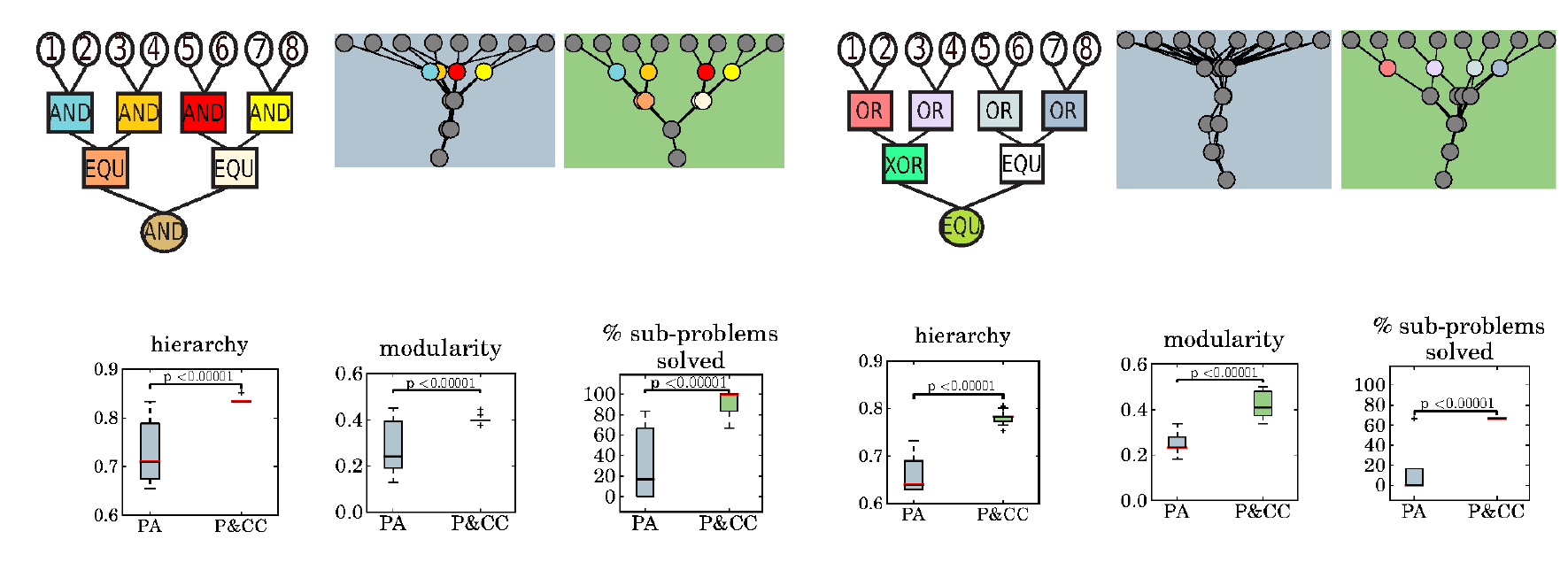}
% \vspace{-10pt}

\caption{Our results are qualitatively unchanged on different problems: P\&CC networks are significantly more modular, hierarchical, and solve more sub-problems than PA networks on different, hierarchical Boolean-logic problems. For each problem, an example evolved network (specifically, the one with median hierarchy) from each treatment is shown. Figs. S2 and S3 show the final, evolved network from each replicate for both treatments on both problems. Note that for the problem on the right, an extra layer of hidden nodes was added due to the complexity of the problem (Methods).}
 \label{fig:generality}
%\vspace{-5.em}
\end{figure*}

To test the generality of our hypothesis that the addition of a connection cost promotes the evolution of hierarchy in addition to modularity, we repeated our experiments with different Boolean-logic problems. For all problem variants, P\&CC networks are significantly more hierarchical, modular, and solve more sub-problems than PA networks (Fig.~\ref{fig:generality}).  The P\&CC treatment also evolved high-performing networks significantly faster on all of these additional problems (Fig. S13). 

\section*{Discussion} The evolution of hierarchy is likely caused by multiple factors. These results suggest that one of those factors is indirect selection to reduce the net cost of network connections.  Adding a cost for connections has previously been shown to evolve modularity~\cite{clune2013:evolutionary,huizingaevolving}; the results in this paper confirm that finding and further show that a cost for connections also leads to the evolution of hierarchy. Moreover, the hierarchy that evolves is functional, in that it involves solving a problem by recursively combining solutions to sub-problems. It is likely that there are other forces that encourage the evolution of hierarchy, and that this connection cost force operates in conjunction with them; identifying other drivers of the evolution of hierarchy and their relative contributions is an interesting area for future research.  

These results also reveal that, like modularity~\cite{Kashtan2005,clune2013:evolutionary}, hierarchy improves evolvability. While modularity and hierarchy are correlated, the experiments we conducted where we explicitly select for networks that are hierarchical, but non-modular, reveal that hierarchy improves evolvability even when modularity is discouraged. 

An additional factor that is present in modular and hierarchal networks is \emph{sparsity}, a term meaning that only a few connections exist of the total that could. It is possible that this property explains some or even all of the evolvability benefits of modular, hierarchical networks. Future work is needed to address the difficult challenge of experimentally teasing apart these related properties.

As has been pointed out for the evolution of modularity~\cite{clune2013:evolutionary}, even if hierarchy, once present, is directly selected for because it increases evolvability, that does not explain its evolutionary origins, because enough hierarchy has to be present in the first place before those evolvability gains can be selected for. 
%Even if hierarchy is ultimately maintained or elevated by selection for evolvability, 
This paper offers one explanation for how sufficient hierarchy can emerge in the first place to then provide evolvability (or other) benefits that can be selected for. 

This paper shows the effect of a connection cost on the evolution of hierarchy via experiments on many variants of one class of problem (hierarchical logic problems with many inputs and one output). In future work it will be interesting to test the generality of these results across different classes of problems, including non-hierarchical problems. The data in this paper suggest that a connection cost will always make it more likely for hierarchy to evolve, but it remains an open, interesting question how wide a range of problems hierarchy will evolve on even with a connection cost.

%A connection cost could thus be a bootstrapping step that enables hierarchy to be further selected for due to other reasons. 
In addition to shedding light on why biological networks evolve to be hierarchical, this work also lends additional support to the hypothesis that a connection cost may also drive the emergence of hierarchy in human-constructed networks, such as company organizations~\cite{R.Guimera:2001}, road systems~\cite{louf2013emergence}, and the Internet~\cite{internethierarchy}.
Furthermore, knowing how to evolve hierarchical networks can improve medical research~\cite{li2014analysis,zhang2014network}, which benefits from more biologically realistic, and thus hierarchical, network models~\cite{shmulevich2002probabilistic,albert2005scale}.

The ability to evolve hierarchy will also aid fields that harness evolution to automatically generate solutions to challenging engineering problems~\cite{keane2006genetic,floreano:2008bio}, as it is known that hierarchy is a beneficial property in engineered designs~\cite{lipson:principles,suh1990principles}. In fact, artificial intelligence researchers have long sought to evolve computational neural models that have the properties of modularity, regularity, and hierarchy~\cite{lipson:principles,stanley2003taxonomy,hornby2005measuring,clune2011:performance,gruau1994automatic}, which are key enablers of intelligence in animal brains~\cite{lipson:principles,nolfi2000evolutionary:Latex,striedter2005principles:LaTeX,wagner:perspective}. It has recently been shown that combining techniques known to produce regularity~\cite{stanley2009hypercube} with a connection cost produces networks that are both modular and regular~\cite{huizingaevolving}. This work suggests that  doing so can produce networks that have all three properties, a hypothesis we will confirm in future work. Being able to create networks with all three properties will both improve our efforts to study the evolution of natural intelligence and accelerate our ability to recreate it artificially.   

\section*{Method and Materials}
\subsection*{Experimental setup}
There were 30 trials per treatment. Each trial is an independent evolutionary process that is initiated with a different random seed, meaning that the sequence of stochastic events that drive evolution (e.g. mutation, selection) are different. Each trial lasted 25000 generations and had a population size of 1000. Unless otherwise stated, all analyses and visualizations are based on the highest-performing network per trial (ties are broken randomly). 

\subsection*{Statistics}
The test of statistical significance is the Mann-Whitney-Wilcoxon rank-sum test, unless otherwise stated. We report and plot medians $ \pm  $ 95\% bootstrapped confidence intervals of the medians, which are calculated by re-sampling the data 5000 times. For visual clarity we reduce the re-sampling noise inherent in bootstrapping by smoothing confidence intervals with a median filter (window size of 101). %, meaning the median spans a total of 2020 generations.

\subsection*{Evolutionary algorithm}
Networks evolve via a multi-objective evolutionary algorithm called the Non-dominated Sorting Genetic Algorithm version II (NSGA-II), which was first introduced in \cite{Deb:2001}. While the original NSGA-II weights all objectives equally, to explore the consequence of having the performance objective be more important than the connection cost objective, Clune et al. \cite{clune2013:evolutionary} created a stochastic version of NSGA-II (called probabilistic NSGA-II, or PNSGA) where each objective is considered for selection with a certain probability (a detailed explanation can be found in \cite{clune2013:evolutionary} and Fig. S15a). Specifically, performance factors into selection 100\% of the time and the connection cost objective factors in $p$ percent of the time. 
Preliminary experiments for this paper demonstrated that values of $p$ of 0.25, 0.5, and 1.0 led to qualitatively similar results. However, for simplicity and because the largest differences between P\&CC and PA treatments resulted from $p=1$, we chose that value as the default for this paper. Note that when {$p=1$}, NSGA-II and PNSGA are identical.

\subsection*{Behavioral diversity}
Evolutionary algorithms notoriously get stuck in local optima (locally, but not globally, high-fitness areas), in part because limited computational resources require smaller population sizes than are found in nature~\cite{floreano:2008bio}. To make computational evolution more representative of natural evolutionary populations, which exhibit more diversity, we adopt the common technique of promoting behavioral diversity~\cite{floreano:2008bio,JBM:2009,JBM:2010diveritymeasure,JBM:2010epericalDiversity,Risi:2009,clune2013:evolutionary} by adding another independent objective that rewards individuals for behaving differently than other individuals in the population. This diversity objective factored into selection 100\% of the time. Preliminary experiments confirmed that this diversity-promoting technique is necessary. Without it, evolution does not reliably produce functional networks in either treatment, as has been previously shown when evolving networks with properties such as modularity~\cite{clune2013:evolutionary,huizingaevolving}.

To calculate the behavioral diversity of a network, for each input we store a network's output (response) in a binary vector; output values {\footnotesize{$>$}} 0 are stored as 1 and 0 otherwise. How different a network's behavior is from the population is calculated by computing the Hamming distance between that network's output vector and all the output vectors of all other networks (and normalizing the result to get a behavioral diversity measure between {\footnotesize{$0$}} and {\footnotesize{$1$}}).

\subsection*{Connection cost}
Following \cite{clune2013:evolutionary}, the connection cost of a network is computed after finding an optimal node placement for internal (hidden) nodes (input and output nodes have fixed locations) that minimizes the overall network connection cost. These locations can be computed exactly~\cite{chklovskii2004exact}. Optimizing the location of internal nodes is biologically motivated; there is evidence that the location of neurons in animal nervous systems are optimized to minimize the summed length of connections between them~\cite{cherniak2004global,chen2006wiring,chklovskii2004exact}. The overall network connection cost is then computed as the summed squared length of all connections. Network visualizations show these optimal neural placements.  

Network nodes are located in a two-dimensional Cartesian space {\footnotesize{($x$,$y$)}}. The locations of input and output nodes are fixed, and the locations of hidden nodes vary according to their optimal location as just described. 
 
For all problems, the inputs have {\footnotesize{$x$}} coordinates of {\footnotesize{$\{-3.5, -2.5, . . . , 2.5, 3.5\}$}}, a {\footnotesize{$y$}} coordinate of {\footnotesize{$(0)$}}, and the output is located at (0,4), except for the problem in Fig. 8, right, which has an output located at {\footnotesize{$(0,5)$}} because of the extra layer of hidden neurons.

\subsection*{Network model and its biological relevance}
The network model is multi-layered and feed-forward, meaning a node at layer {\small{$n$}} receives incoming connections only from nodes at layer {\small{$n-1$}} and has outgoing connections only to nodes at layer {\small{$n+1$}}. This network model is common for investigating questions in systems biology~\cite{alon2007:introduction,Kashtan2007,karlebach2008modelling}, including studies into the evolution of modularity~\cite{clune2013:evolutionary,kashtan2005spontaneous}. While the layered and feed-forward nature of networks may contribute to elevated hierarchy, this network architecture is the same across all treatments and we are interested in the \textit{differences} between levels of hierarchy that occur with and without a connection cost. 

For the main problem, AND-XOR-AND, networks are of the form {\footnotesize{$8\setminus4\setminus4\setminus2\setminus1$}}, which means that there are 8 input nodes, 3 hidden layers each having 4, 4 and 2 nodes respectively, and 1 output node. The integers {\footnotesize{\{-2, -1, 1, 2\}}} are the possible values for connection weights, whereas the possible values for the biases are {\footnotesize{$\{-2, -1, 0, 1, 2\}$}}. 

Information flows through the network in discrete time stpdf one layer at a time. The output {\footnotesize{$y_{j}$}} of node {\footnotesize{$j$}} is the result of the function:
 \begin{eqnarray}
\footnotesize{y_{j} =  \textstyle {\tanh \left ( \lambda \left (\sum_{i \in I_{j} }^{}\omega _{ij}y_{i}+b_{j}  \right ) \right )}}
\end{eqnarray}  

where {\footnotesize{{$I_{j}$}}} is the set of nodes connected to node {\footnotesize{$j$}}, {\footnotesize{$w _{ij}$}} is the connection strength between node {\footnotesize{$i$}} and node {\footnotesize{$j$}},  {\footnotesize{$y_{i}$}} is the output value of node {\footnotesize{$i$}}, and {\footnotesize{$b_{j}$}} is a bias. The bias determines at which input value the output changes from negative to positive. The function, {\footnotesize{$\tanh(x)$}}, is an activation function that guarantees that the output of a node is in the range of {\footnotesize{$[-1,1]$}}. The slope of the transition between the two extreme output values is determined by {\footnotesize{$\lambda$}}, which is here set to 20 (Figs. S15B-C).

Following (16), in all treatments, the initial number of connections in networks is randomly chosen between 20 and 100. If that chosen number of initial connections is more than the maximum number of connections that a network can have (which is 58), the network is fully connected. When random, valid networks are generated, the initial number of connections ranges from the minimum number needed, which is 11, to the maximum number possible, which is 58. Each connection is assigned a random weight selected from the possible values for connection weights and is placed between randomly chosen, unconnected neurons residing on two consecutive layers. Our results are qualitatively unchanged when the initial number of connections is smaller than the default, i.e. when evolution starts with sparse networks (Fig. S14).

\subsection*{Mutations}
To create offspring, a parent network is copied and then randomly mutated. Each network has a {\footnotesize{$20\%$}} chance of having a single connection added. Candidates for new connections are pairs of unconnected nodes that reside on two consecutive layers. Each network also has a {\footnotesize{$20\%$}} chance of a randomly chosen connection being removed. Each node in the network has {\footnotesize{$ 0.067 \%$}} chance of its bias being incremented or decremented with both options equally probable; five values are available {\footnotesize{$\{-2, -1, 0, 1, 2\}$}} and mutations that produce values higher or lower than these values are ignored. Each connection in the network has {\footnotesize{$\frac{2.0}{n}$}} chance of its weight being incremented or decremented, where {\footnotesize{$n$}} is the total number of connections in the network. Because weights must be in the set of possible values {\footnotesize{$\{-2, -1, 1, 2\}$}}, mutations that produce values higher or lower than these four values are ignored.

\subsection*{Modularity}Network modularity is calculated with an efficient approximation~\cite{leicht2008community} of the commonly used Q metric developed by \cite{newman2006modularity}. Because that metric has been extensively described before \cite{newman2006modularity}, here we only describe it briefly. The Q metric defines network modularity, for a particular division of the network, as the number of within-module edges minus the expected number of these edges in an equivalent network, where edges are placed randomly \cite{newman2006modularity}. The formula for this metric is: 
\begin{equation}
\footnotesize{Q = \frac{1}{m}\sum_{ij}[A_{ij} - \frac{k_{i}^{in}k_{j}^{out}}{m}]\sigma_{ci,cj} }
\end{equation}
Where {\footnotesize{$k_{i}^{in}$}} and {\footnotesize{$k_{j}^{out}$}} are the in- and out-degree of node {\footnotesize{$i$}} and {\footnotesize{$j$}}, respectively,  {\footnotesize{$m$}} is the total number of edges in the network, {\footnotesize{$A_{ij}$}} is the connectivity matrix, which is 1 if there is an edge from node {\footnotesize{$i$}} to {\footnotesize{$j$}}, and 0 otherwise, and {\footnotesize{$\sigma_{ci,cj}$}} is a function whose value is 1 if nodes {\footnotesize{$i$}} and {\footnotesize{$j$}} belong to the same module, and {\footnotesize{$0$}} otherwise.

\subsection*{Hierarchy}
Our hierarchy measure comes from~\cite{mones2012:hierarchy}. It ranks nodes based on their influence. A node's influence on a network equals the portion of a network that is reachable from that node (respecting the fact that edges are directed). Based on this metric, the larger the proportion of a network a node can reach, via its outgoing edges, the more influential it is. For example, a root node has more influence because a path can be traced from it to every node in the network, whereas leaf nodes have no influence. The metric calculates network hierarchy by computing the \textit{heterogeneity} of the influence values of all nodes in the network. Intuitively, node-influence heterogeneity is high in hierarchical networks (where some nodes have a great deal of influence and others none), and low in non-hierarchical networks (e.g. in a fully connected network the influence of nodes is perfectly homogeneous). 

Because of the non-linear function that maps a node's inputs to its output, even a small change in the input to a node can change whether it fires. For that reason, it is difficult to determine the influence one node has on another based on the strength of the connection between them. We thus calculate hierarchy scores by looking only at the presence of connections between nodes, ignoring the strength of those connections. The score for a weighted directed network is calculated by:

\begin{equation}
\footnotesize{\frac{\sum_{i\in V}[C_{R}^{max} - C_{R}(i)]}{N-1}}
\end{equation}
{\footnotesize{$C_{R}^{max}$}} is the highest influence value and {\footnotesize{$V$}} represents a set of all nodes in the network. {\footnotesize{$N$}} is the number of nodes in the network. Each node in the network is represented by {\footnotesize{$i$}}. {\footnotesize{$C_{R}(i)$}}, the influence value of node {\footnotesize{$i$}}, is given by:
\begin{equation}
\footnotesize{C_{R}(i) = \frac{1}{N-1} \sum_{j:0<d^{out}(i,j)<\infty}[\frac{1}{d^{out}(i,j)}]}
\end{equation}
Here, {\footnotesize{$d^{out}(i,j)$}} is the length of the path that goes from node {\footnotesize{$i$}} to node {\footnotesize{$j$}}, meaning it is the number of outgoing connections along the path. 

\subsection*{Functional hierarchy}  
As a proxy for quantifying functional hierarchy, we measure the percent of logic sub-problems of an overall problem that are solved by part of a network. Note that an overall logic problem can be solved without solving any specific sub-problem (e.g. in the extreme, if the entire problem is computed in one step). We determine whether a logic sub-problem is solved as follows: for all possible inputs to a network, a neuron solves a logic gate (a sub-problem) if, for its outputs, there exists any threshold value that correctly separates all the True answers from the False answers for the logic gate in question. We also consider a sub-problem as solved if it is solved by a \textit{group} of neurons on the same layer. To check for this case, we consider all possible groupings of neurons in a layer (groups of all sizes are checked). We sum the outputs of the neurons in a group and see if there is a threshold that correctly separates True and False for the logic sub-problem on all possible network inputs. To prevent counting solutions multiple times, each sub-problem is considered only once: i.e. the algorithm stops searching when a sub-problem is found to be solved.

\section*{Supplementary Information}
%You can find supplementary information from this link:\\
%https://github.com/henokyen/arxiv\_si/blob/master/arxiv\_si.pdf
%\href{https://github.com/henokyen/arxiv_si/blob/master/arxiv_si.pdf}{Supplementary Information} 

\renewcommand{\figurename}{Fig. S}
\setcounter{figure}{0}
\setlength{\abovecaptionskip}{15pt plus 3pt minus 2pt}
\clearpage
%S1
\begin{figure*}
\vspace{-10pt}

   {\setlength{\tabcolsep}{0em}
     \begin{tabular}{ p{0.3\textwidth}p{.7\textwidth}}  
     \vspace{-43pt}\hspace{140pt}A. &         
     \vspace{-47pt}\hspace{10pt}\includegraphics[width=.33\textwidth]{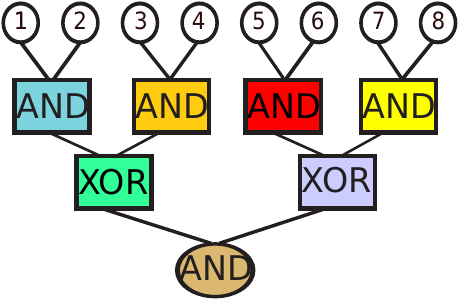}     
      \end{tabular}
    }
\end{figure*}
\begin{figure*}[htb!]
\centering
\begin{minipage}[t]{0.49\textwidth}
\centering
   {\setlength{\tabcolsep}{0em}
      \begin{tabular}{p{0.5\textwidth} p{1.4\textwidth}}           
       \hspace{-100pt} B. Performance Alone (PA) &
        \hspace{20pt} C. Performance \& Connection Cost (P\&CC) 
      \end{tabular}
      \vspace{-15pt}
      \begin{tabular}{p{1.0\textwidth} p{1.0\textwidth}}           
      \hspace{-97pt}\includegraphics[width=.84\textwidth]{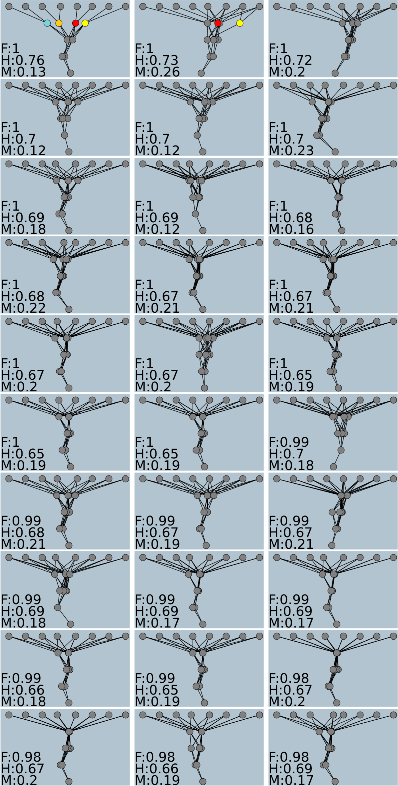}&
      \hspace{-97pt} \includegraphics[width=.84\textwidth]{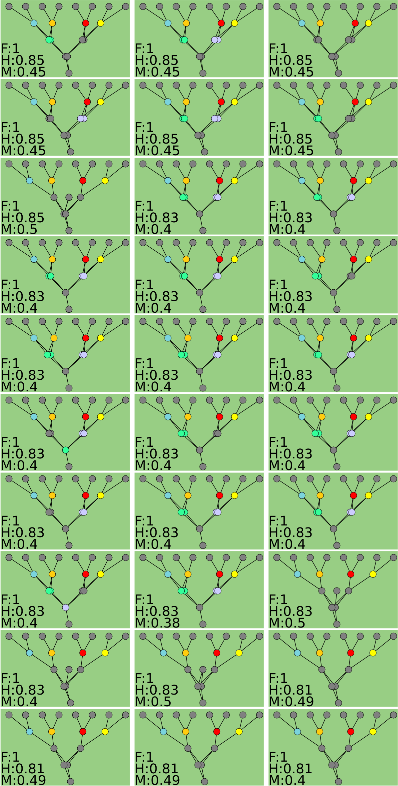}
      \end{tabular}
    }
\end{minipage}
\caption{The addition of a connection cost leads to the evolution of hierarchical, modular, and functionally hierarchical networks. In this visualization, networks are first sorted by fitness (F), then by hierarchy (H), and finally by modularity (M). Network nodes are colored if they solve a logic subproblem of the overall problem (Methods). \textbf{(A)} The main experimental problem in the paper, AND-XOR-AND. \textbf{(B-C)} The highest-performing networks at the end of each trial of the performance alone (PA) treatment (left) are less hierarchical, modular, and functionally hierarchical than networks from the performance and connection cost (P\&CC) treatment (right).}
%Network information from the main experiment
\label{fig:S1}
%\vspace{.02em}
\end{figure*}
\pagebreak
%S2
\begin{figure*}
   {\setlength{\tabcolsep}{0em}
   \centering
     \begin{tabular}{ p{0.3\textwidth}p{0.9\textwidth}}  
     \vspace{-43pt}\hspace{140pt}A. &         
     \vspace{-47pt}\hspace{10pt}\includegraphics[width=0.33\textwidth]{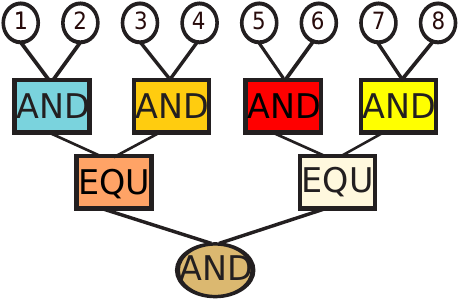}     
      \end{tabular}
    }
\end{figure*}
\begin{figure*}[htb!]
\centering
\begin{minipage}[t]{0.49\textwidth}
\centering
   {\setlength{\tabcolsep}{0em}
      \begin{tabular}{p{0.5\textwidth} p{1.1\textwidth}}           
       \hspace{-100pt} B. Performance Alone (PA) &
        \hspace{20pt} C. Performance \& Connection Cost (P\&CC) 
      \end{tabular}
      \vspace{-15pt}
      \begin{tabular}{p{1.0\textwidth} p{1.0\textwidth}}           
      \hspace{-97pt}\includegraphics[width=.84\textwidth]{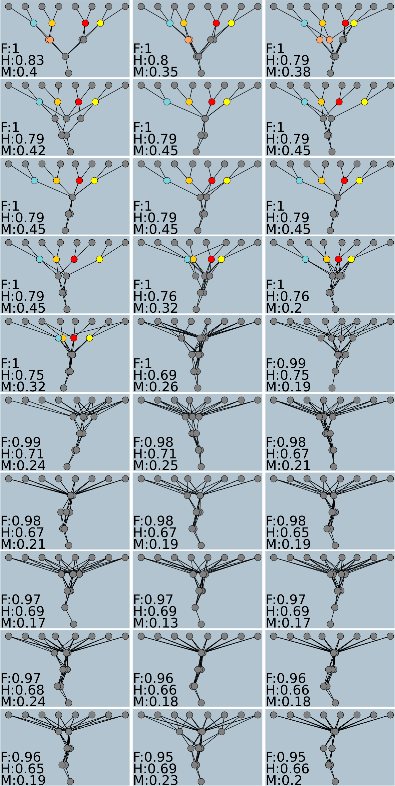}&
      \hspace{-97pt} \includegraphics[width=.84\textwidth]{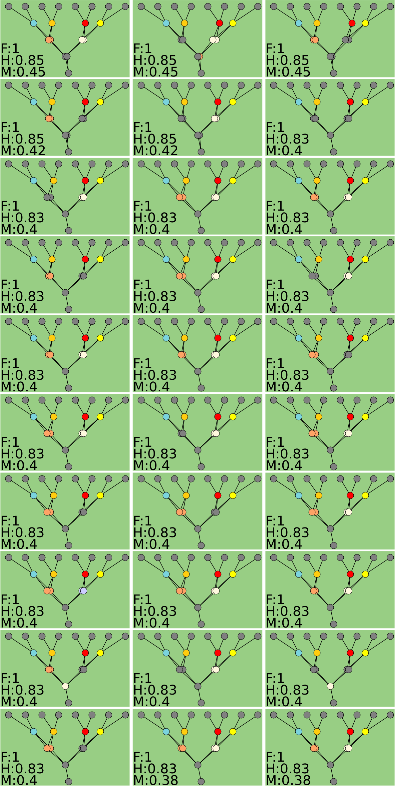}
      \end{tabular}
    }
\end{minipage}
\caption{The results from the main experiment are qualitatively the same on a second, different, hierarchical  problem: AND-EQU-AND \textbf{(A)}.  The highest-performing networks at the end of each trial of the performance alone (PA) treatment \textbf{(B)} are less hierarchical, modular, and functionally hierarchal than networks from the performance and connection cost (P\&CC) treatment \textbf{(C)}. Networks are first sorted by fitness (F), then by hierarchy(H), and finally by modularity (M).
}
\label{fig:S2}
\end{figure*}
\pagebreak
%S3
\begin{figure}
\vspace{-20pt}
   {\setlength{\tabcolsep}{0em}
   \centering
     \begin{tabular}{ p{0.3\textwidth}p{0.7\textwidth}}  
      \vspace{-43pt}\hspace{140pt}A. &         
      \vspace{-47pt}\hspace{10pt}\includegraphics[width=0.265\textwidth]{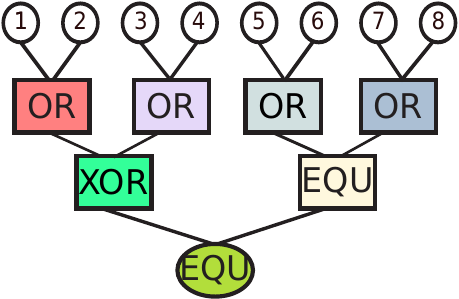}     
      \end{tabular}
    }
  \end{figure}
\begin{figure*}[htb!]
\centering
\begin{minipage}[t]{0.545\textwidth}
\centering
\vspace{-35pt}
   {\setlength{\tabcolsep}{0em}
      \begin{tabular}{p{0.6\textwidth} p{1.1\textwidth}}           
      \vspace{75pt}\hspace{-85pt} B. Performance Alone (PA) &
      \vspace{75pt}\hspace{2pt} C. Performance \& Connection Cost (P\&CC) 
      \end{tabular}
      \begin{tabular}{p{1.0\textwidth} p{1.0\textwidth}}           
      \hspace{-80pt}\includegraphics[width=.61\textwidth]{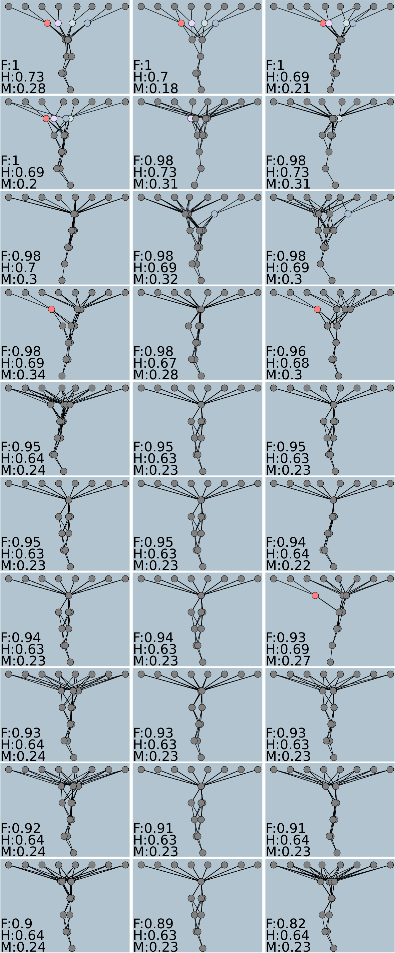}&
      \hspace{-97pt}\includegraphics[width=.61\textwidth]{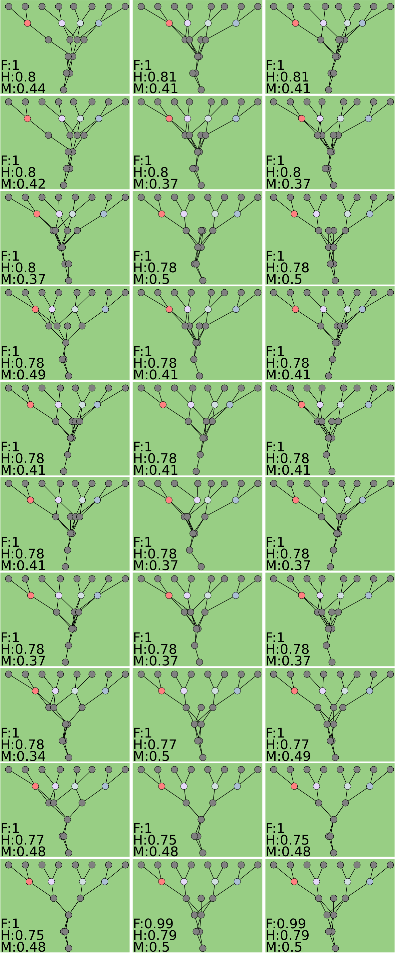}
      \end{tabular}
    }
\end{minipage}
\caption{ The results from the main experiment are qualitatively the same on a third, different, hierarchical problem: OR-XOR/EQU-EQU. See the previous caption for a lengthier explanation. These networks have an extra layer of hidden nodes vs. the default network model owing to the extra complexity of the last logic gate, EQU.}
\label{fig:S3}
\end{figure*}
\pagebreak
%S4
\begin{figure*}
\vspace{-30pt}
   {\setlength{\tabcolsep}{0em}
     \begin{tabular}{ p{0.3\textwidth}p{0.7\textwidth}}  
     \vspace{-43pt}\hspace{140pt}A. &         
     \vspace{-45pt}\hspace{10pt}\includegraphics[width=.3\textwidth]{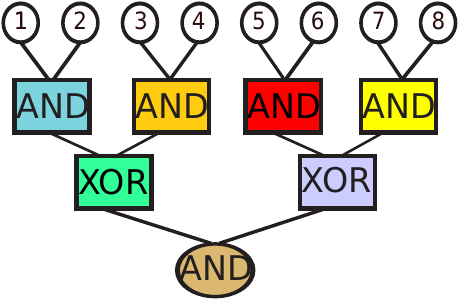}     
      \end{tabular}
    }
\end{figure*}
\begin{figure*}[htb!]
\centering
\begin{minipage}[t]{0.6\textwidth}
\centering
   {\setlength{\tabcolsep}{0em}
      \begin{tabular}{p{0.9\textwidth}}           
      \vspace{-3pt}\hspace{5pt}B. P\&CC-NonMod  
      \end{tabular}
      \centering
      \begin{tabular}{ p{4\textwidth}}           
      \vspace{-0pt}\hspace{20pt}\includegraphics[trim= 1.5cm 7.5cm 9.5cm 0.2cm,clip,scale=0.72]{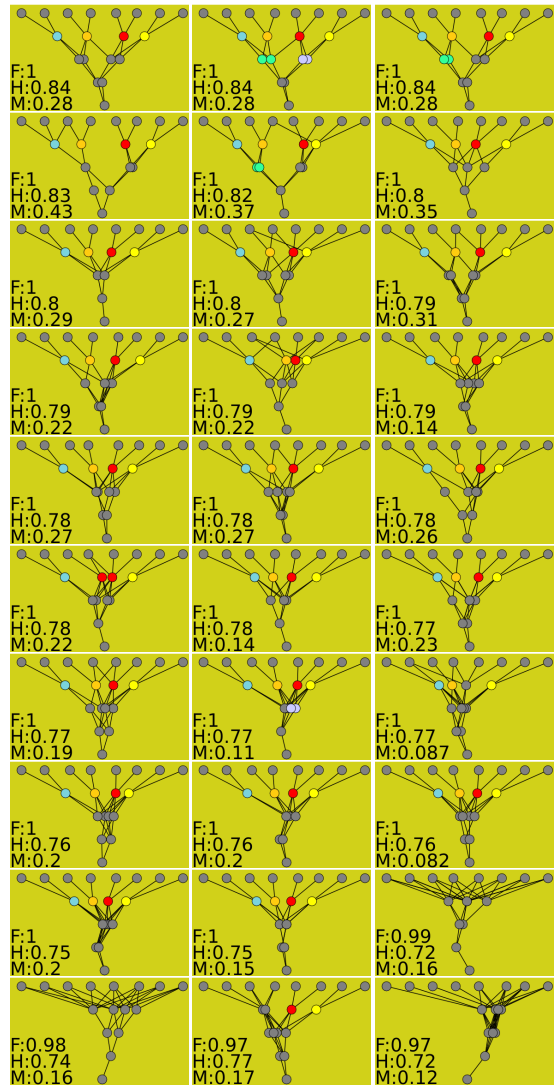}   
      \end{tabular}
    }
\end{minipage}
\caption{Evolving networks with a connection cost, but an additional explicit pressure to be non-modular, produces networks that are hierarchical, but non-modular. These results show that a connection cost promotes hierarchy independent of the modularity-inducing effects of a connection cost. \textbf{(A)} The problem for this experiment, which was the default experiment for the paper (AND-XOR-AND). \textbf{(B)} Almost all of the end-of-run networks from this P\&CC-NonMod treatment are hierarchical, yet have low modularity.} 
\label{fig:S4}
\end{figure*}

\pagebreak
%S5
\begin{figure*}
\vspace{-50pt}
   {\setlength{\tabcolsep}{0em}
     \begin{tabular}{ p{0.4\textwidth}p{.7\textwidth}}  
     \vspace{-13pt}\hspace{140pt}A. &         
     \vspace{-15pt}\hspace{-40pt}\includegraphics[scale=0.19]{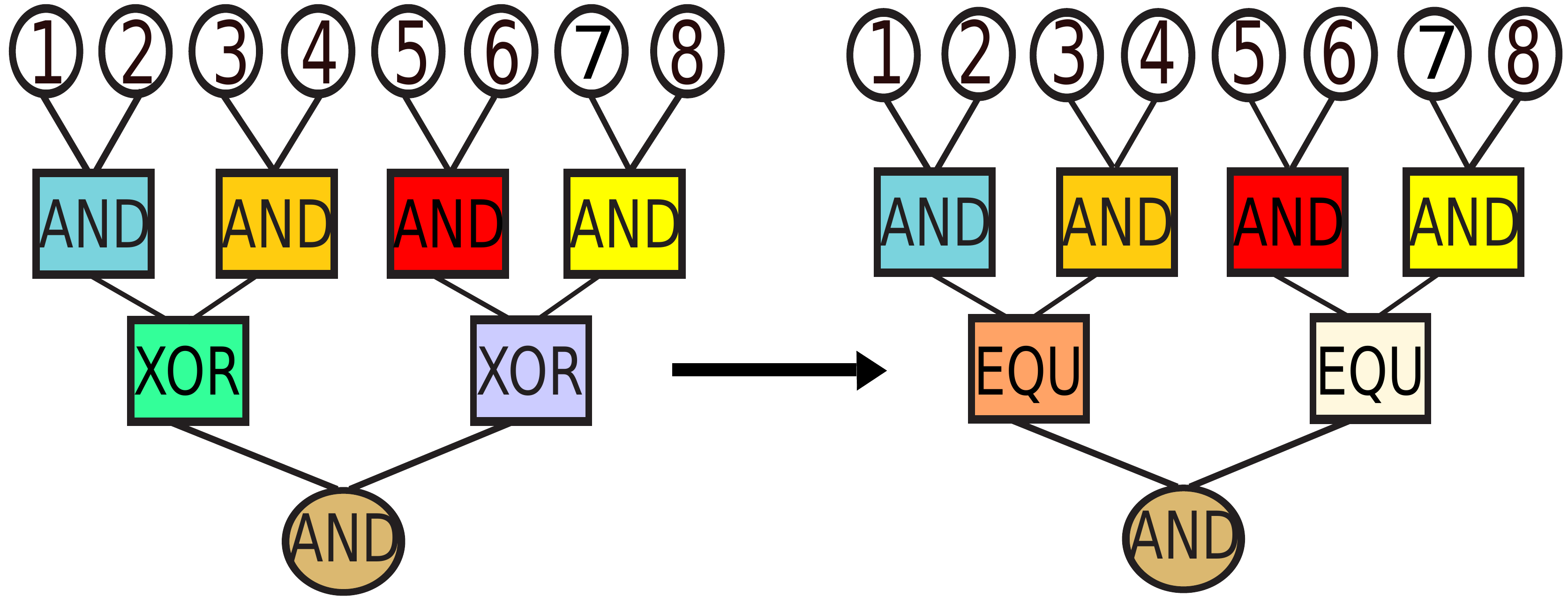}     
      \end{tabular}
    }
  \end{figure*}
  \begin{figure*}[htb!]
\vspace{-30pt}
  \begin{minipage}[t]{0.98\textwidth}
     {\setlength{\tabcolsep}{0em}
        \begin{tabular}{ p{0.9\textwidth}}           
          \vspace{50pt}\hspace{10pt}B. Performance Alone (PA) 
        \end{tabular}
        \centering          
        \includegraphics[scale=0.65]{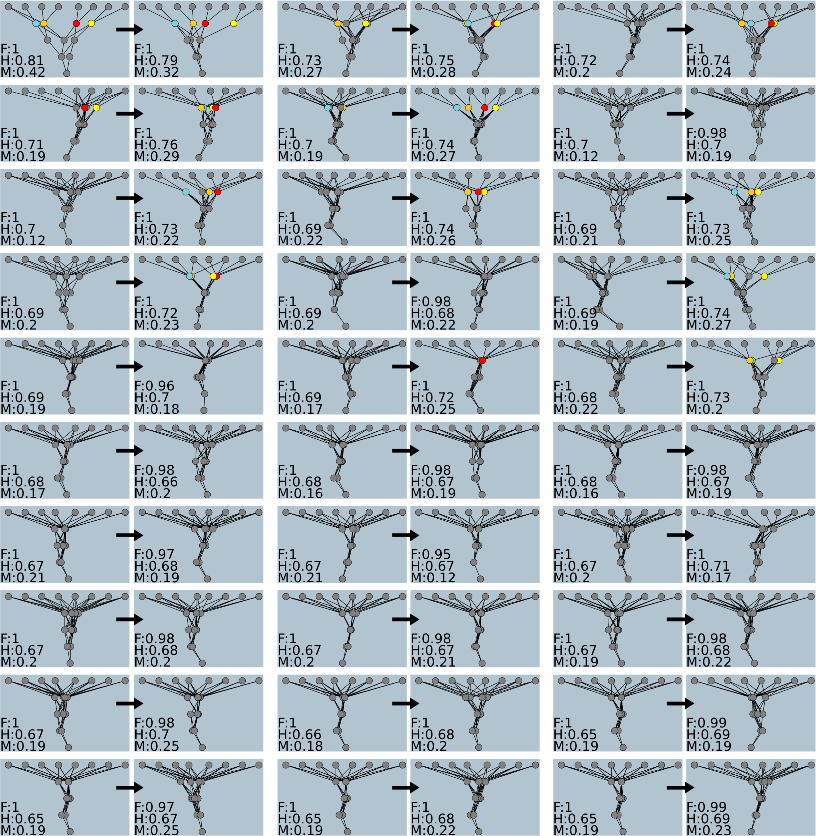}    
     
      }
  \end{minipage}

  \caption{(part 1 of 2) The networks from the PA treatment for the first evolvability experiment, in which networks are first evolved to perfect fitness on the AND-XOR-AND problem and then are transferred (black arrow) to the AND-EQU-AND problem~\textbf{(A)}. The highest-performing network from each replicate in the base environment seeds 30 independent runs in the target environment, leading to a total of 900 replicates per treatment in the target environment.~\textbf{(B)} In this visualization the best-performing networks from the original environment are on the left side of each arrow and on the right side is an example descendant network from the target environment (specifically, the network with median hierarchy).
  }

\label{fig:S5}
  \end{figure*}
\pagebreak
%S6
\begin{figure*}
\centering
  \begin{minipage}[t]{0.49\textwidth}
      {\setlength{\tabcolsep}{0em}
      \centering
      \begin{tabular}{ p{0.55\textwidth}}           
          \vspace{10pt}\hspace{-100pt}Performance \& Connection Cost (P\&CC)      
       \end{tabular}
       \begin{tabular}{ p{4\textwidth}}          
       %   \vspace{-10pt}\hspace{290pt}\includegraphics[trim= 0.1cm 4.5cm 0cm 1.7cm,clip,scale=0.74]{SI/maintoandequalsandPCC.pdf}
           \vspace{-10pt}\hspace{-100pt}\includegraphics[scale=0.74]{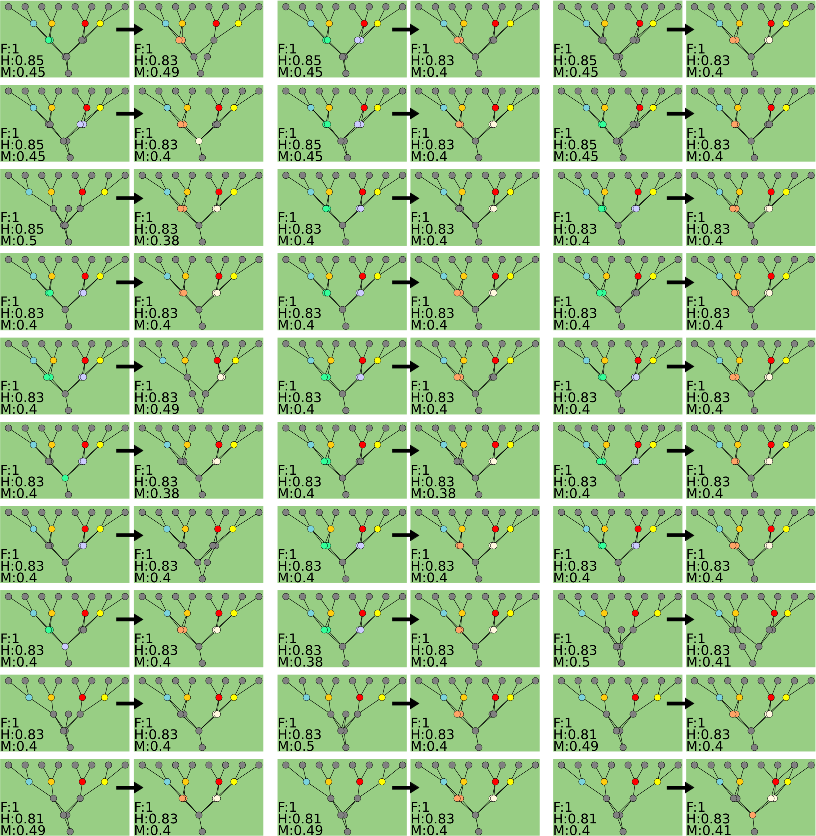}     
       \end{tabular}
      }
  \end{minipage}
    \caption{(part 2 of 2) The networks from the P\&CC treatment for the first evolvability experiment. See the previous caption for a more detailed explanation.}
    \label{fig:S6}
\end{figure*}  
\pagebreak
%S7
\begin{figure*}
\vspace{-3pt}
   {\setlength{\tabcolsep}{0em}
   \centering
     \begin{tabular}{ p{0.4\textwidth}p{.6\textwidth}}  
      \vspace{-27pt}\hspace{100pt}A. &         
      \vspace{-30pt}\hspace{-80pt}\includegraphics[scale=0.26]{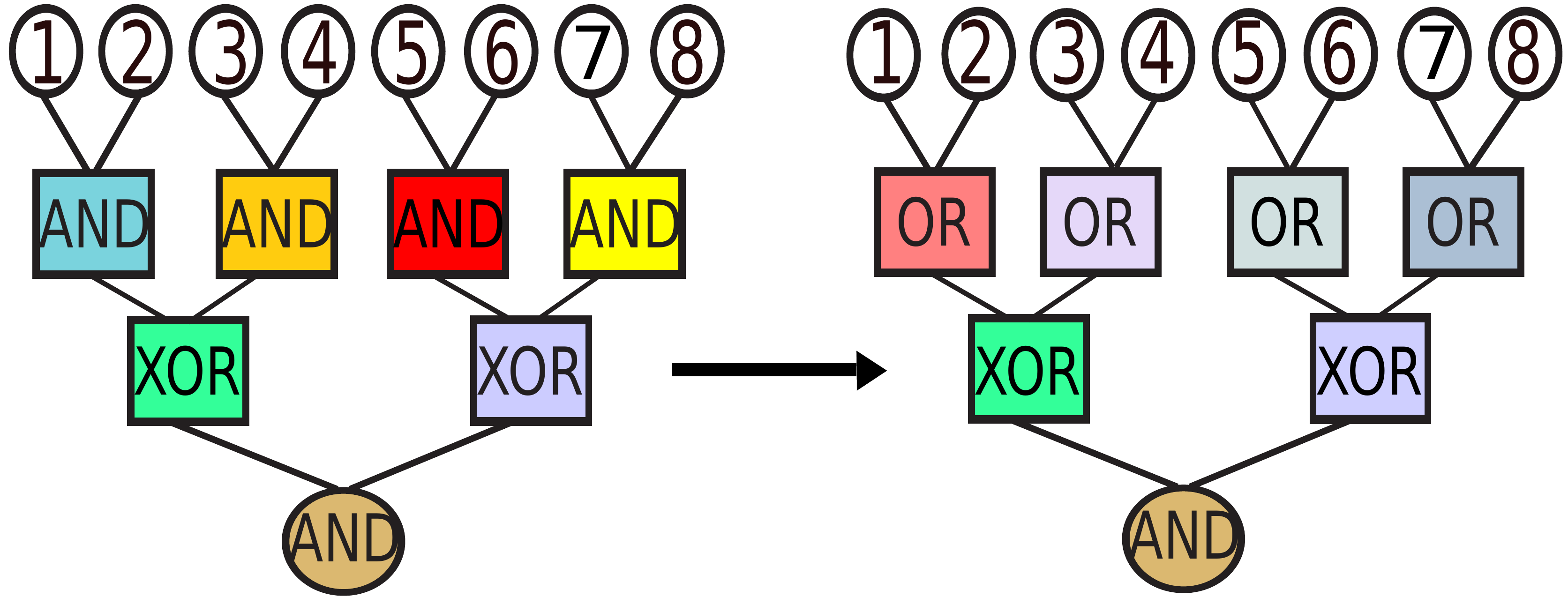}     
      \end{tabular}
    }
  \end{figure*}
  \begin{figure*}
  \centering
  \begin{minipage}[t]{0.4\textwidth}
     {\setlength{\tabcolsep}{0em}
        \begin{tabular}{ p{0.55\textwidth}}           
         \vspace{-7pt}\hspace{-110pt}B. Performance Alone (PA) 
        \end{tabular}
        \begin{tabular}{ p{6\textwidth}}     
             
   %     \vspace{-20pt}\hspace{290pt}\includegraphics[trim= 0.1cm 4.5cm 0cm 1.7cm,clip,scale=0.74]{SI/maintoorxorandPA.pdf}   
        \hspace{-110pt}\includegraphics[scale=0.7]{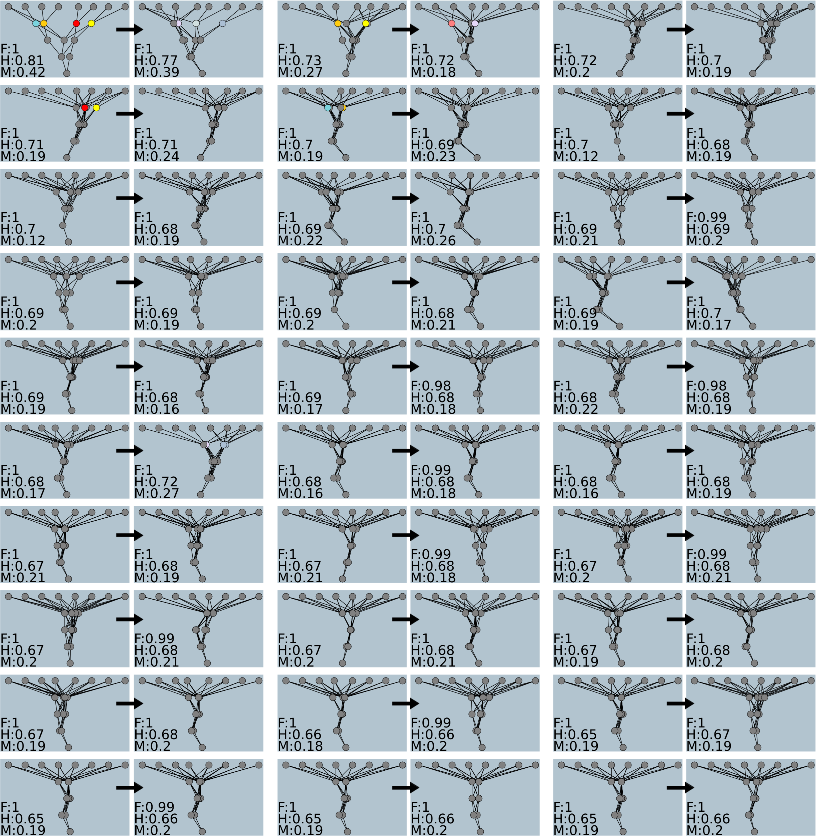} 
        \end{tabular}
      }
  \end{minipage}

  \caption{(part 1 of 2) The second, AND-XOR-AND to OR-XOR-AND, evolvability experiment~\textbf{(A)} and the networks from the PA treatment for this experiment~\textbf{(B)}. Except for a different target environment, this experiment has the same setup as the evolvability experiment in Fig. S\ref{fig:S5}.}
  \label{fig:S7}
  \end{figure*}
\pagebreak 
 %S8
\begin{figure*}
\centering
\begin{minipage}[t]{0.49\textwidth}
   {\setlength{\tabcolsep}{0em}
    \begin{tabular}{ p{0.55\textwidth}}           
        \vspace{-17pt}\hspace{-102pt}Performance \& Connection Cost (P\&CC)      
     \end{tabular}
     \begin{tabular}{ p{4\textwidth}}          
  %   \vspace{-20pt}\hspace{280pt}\includegraphics[trim= 0.1cm 4.5cm 0cm 1.7cm,clip,scale=0.74]{SI/maintoorxorandPCC.pdf}   
     \vspace{-20pt}\hspace{-100pt}\includegraphics[scale=0.74]{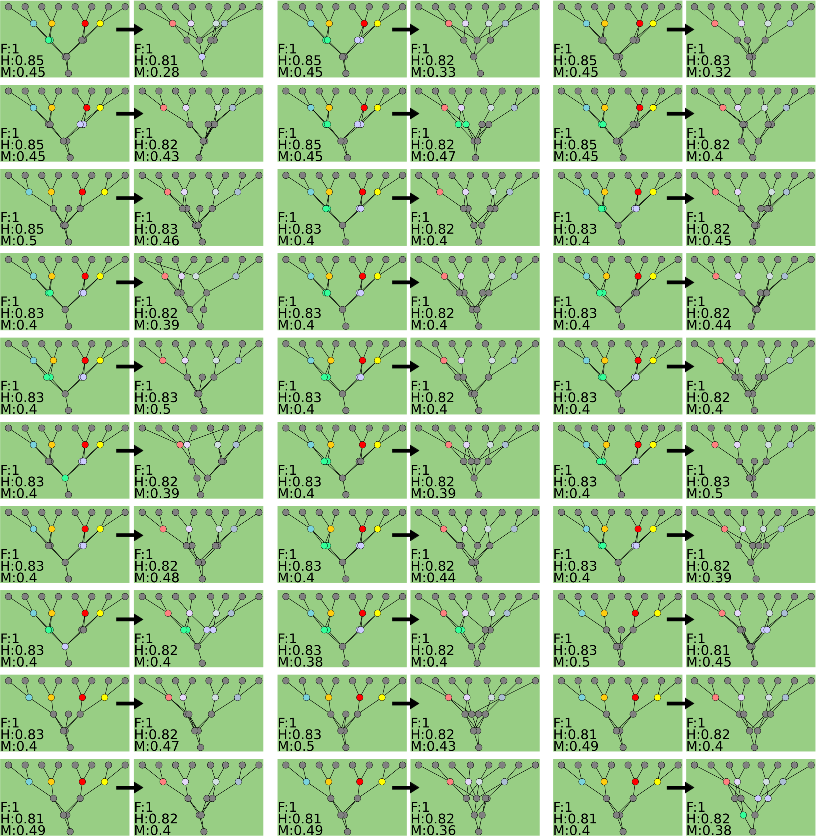}
     \end{tabular}
    }
\end{minipage}
 \caption{(part 2 of 2) The P\&CC treatment networks from the second, AND-XOR-AND to OR-XOR-AND, evolvability experiment (pictured in Fig. S7). Except for a different target environment, this experiment has the same setup as the evolvability experiment shown in Fig. S\ref{fig:S5}.}
 \label{fig:S8}
\end{figure*}
\pagebreak
%S9
\begin{figure*}
\vspace{-17pt}
   {\setlength{\tabcolsep}{0em}
   \centering
     \begin{tabular}{p{0.3\textwidth}p{0.4\textwidth}}  
     \vspace{3pt}\hspace{90pt}A. &         
     \vspace{0pt}\hspace{-35pt}\includegraphics[scale=0.22]{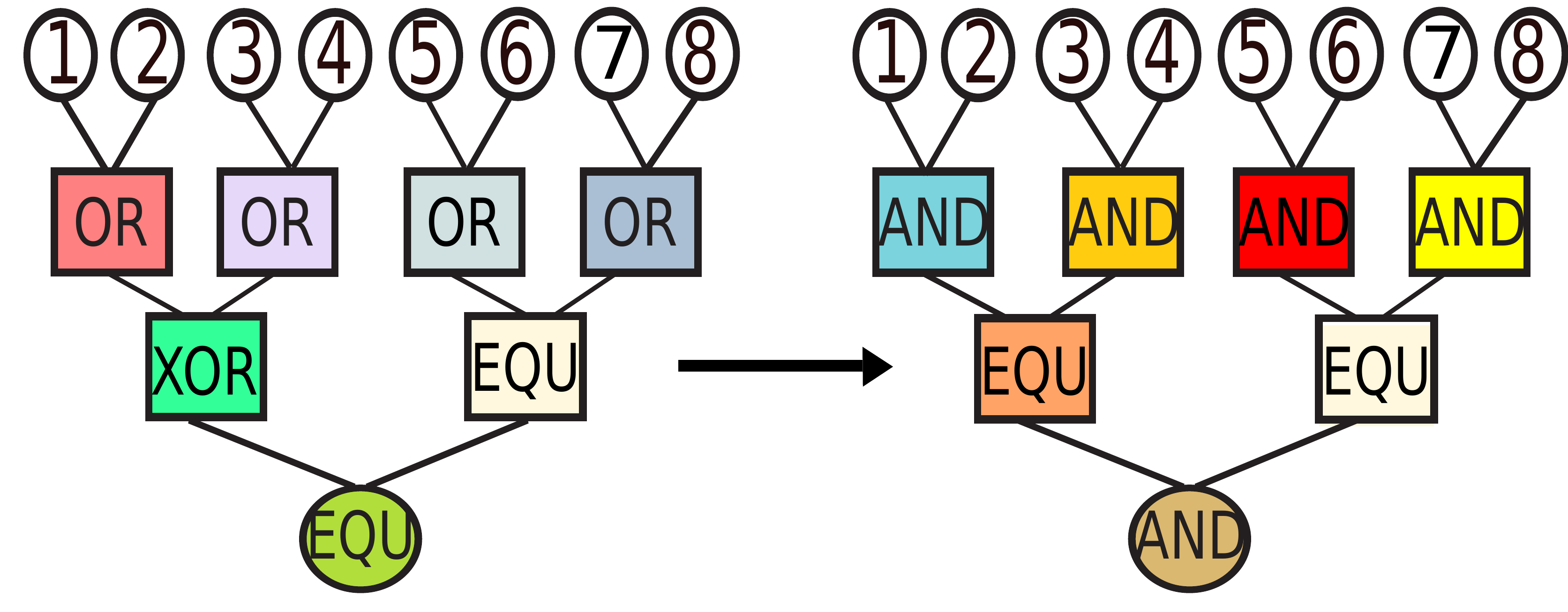}     
      \end{tabular}
    }
\end{figure*}
\begin{figure*}
\centering
\begin{minipage}[t]{0.49\textwidth}
   {\setlength{\tabcolsep}{0em}
      \begin{tabular}{ p{0.55\textwidth}}           
      \vspace{20pt}\hspace{-59pt}B. Performance Alone (PA) 
      \end{tabular}
      \begin{tabular}{ p{4\textwidth}}           
%       \vspace{-10pt}\hspace{295pt}\includegraphics[trim= 0.1cm .7cm 0cm 1.7cm,clip,scale=0.677]{SI/comb_to_andequalsandPA.pdf}   
      \vspace{-10pt}\hspace{-60pt}\includegraphics[scale=0.56]{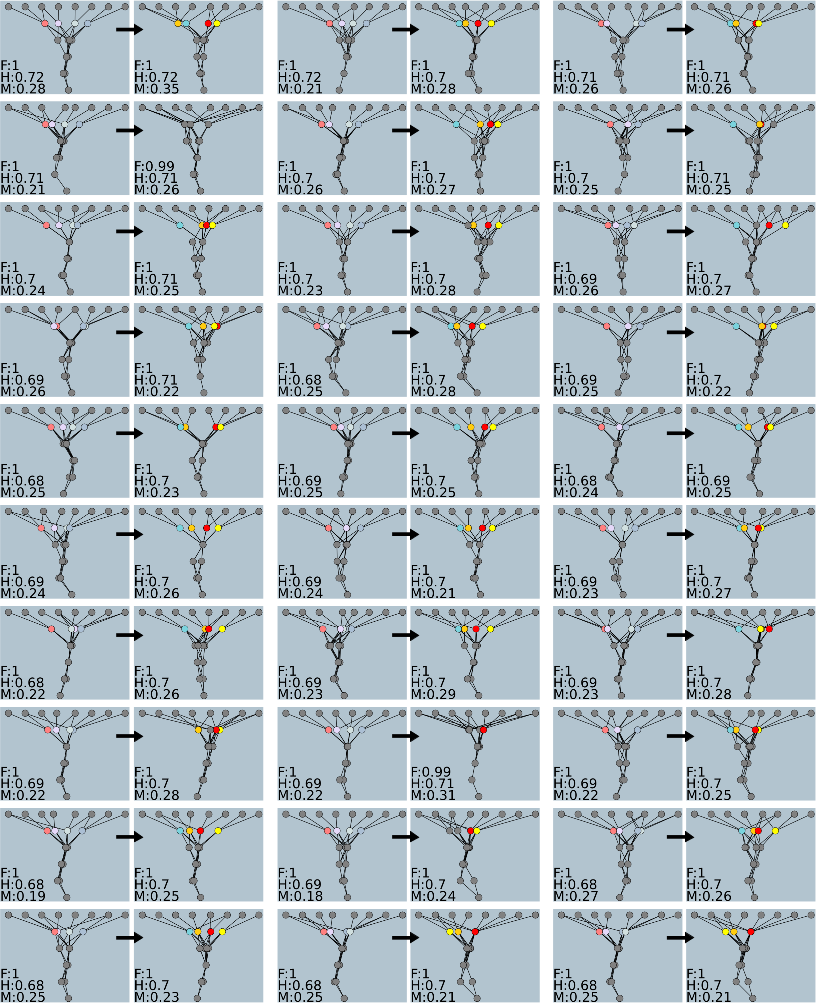}   
      \end{tabular}
    }
\end{minipage}

\caption{(part 1 of 2) The third evolvability experiment, OR-XOR/EQU-EQU to AND-EQU-AND~\textbf{(A)}, and the networks from the PA treatment for this experiment~\textbf{(B)}. Except for a different base environment, this experiment has the same setup as the evolvability experiment shown in Fig. S\ref{fig:S5}.}
\label{fig:S9}
\end{figure*}
\pagebreak
%S10
\begin{figure*}
\centering
 \begin{minipage}[t]{0.49\textwidth}
   {\setlength{\tabcolsep}{0em}
    \begin{tabular}{ p{0.55\textwidth}}           
    \vspace{-50pt}\hspace{-102pt}Performance \& Connection Cost (P\&CC)       
   \end{tabular}
   \begin{tabular}{ p{4\textwidth}}          
   %\vspace{-45pt}\hspace{280pt}\includegraphics[trim= 0cm 0.2cm 0cm 0.2cm,clip,scale=0.74]{SI/comb_to_andequalsandPCC.pdf}   
    \vspace{-45pt}\hspace{-100pt}\includegraphics[scale=0.74]{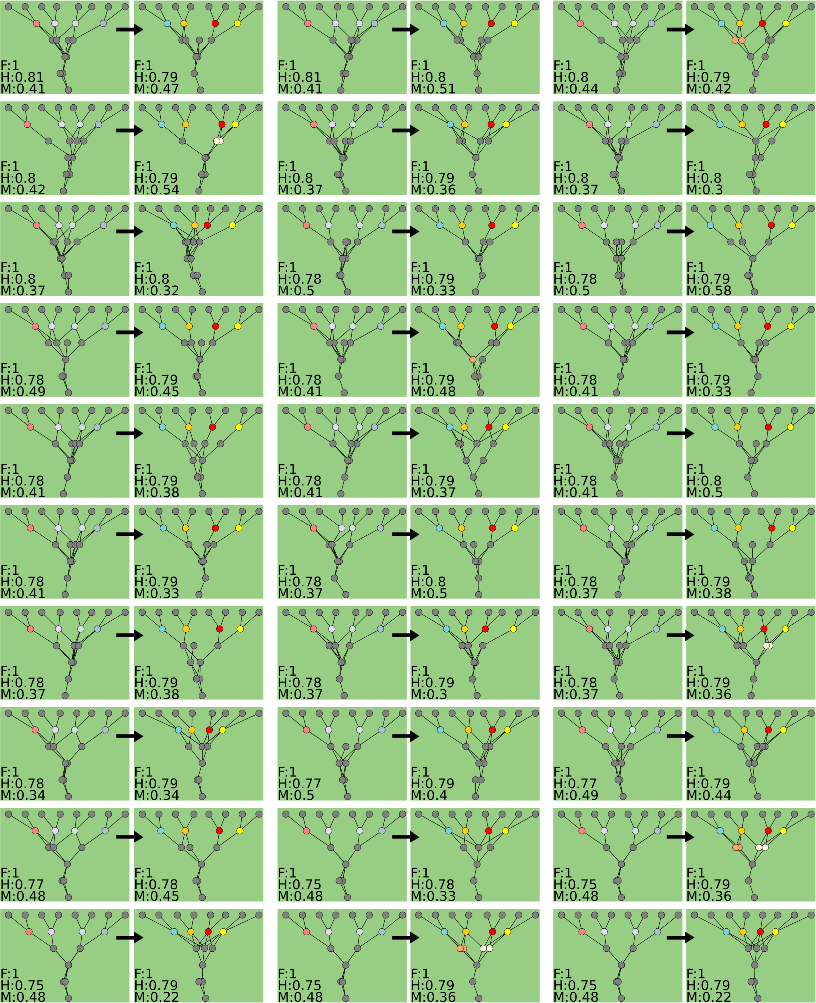} 
    \end{tabular}
  }
    \end{minipage}
    \caption{(part 2 of 2) The networks from the P\&CC treatment for the third, OR-XOR/EQU-EQU to AND-EQU-AND, evolvability experiment (pictured in Fig. S9). Except for a different base environment, this experiment has the same setup as the evolvability experiment in Fig. S\ref{fig:S5}.}
    \label{fig:S10}
\end{figure*}
\pagebreak
%S11
\begin{figure*}
\vspace{-50pt}
   {\setlength{\tabcolsep}{0em}
   \centering   
     \begin{tabular}{ p{0.3\textwidth}p{0.4\textwidth}}  
     \vspace{3pt}\hspace{86pt}A. &         
      \vspace{0pt}\hspace{-35pt}\includegraphics[scale=0.239]{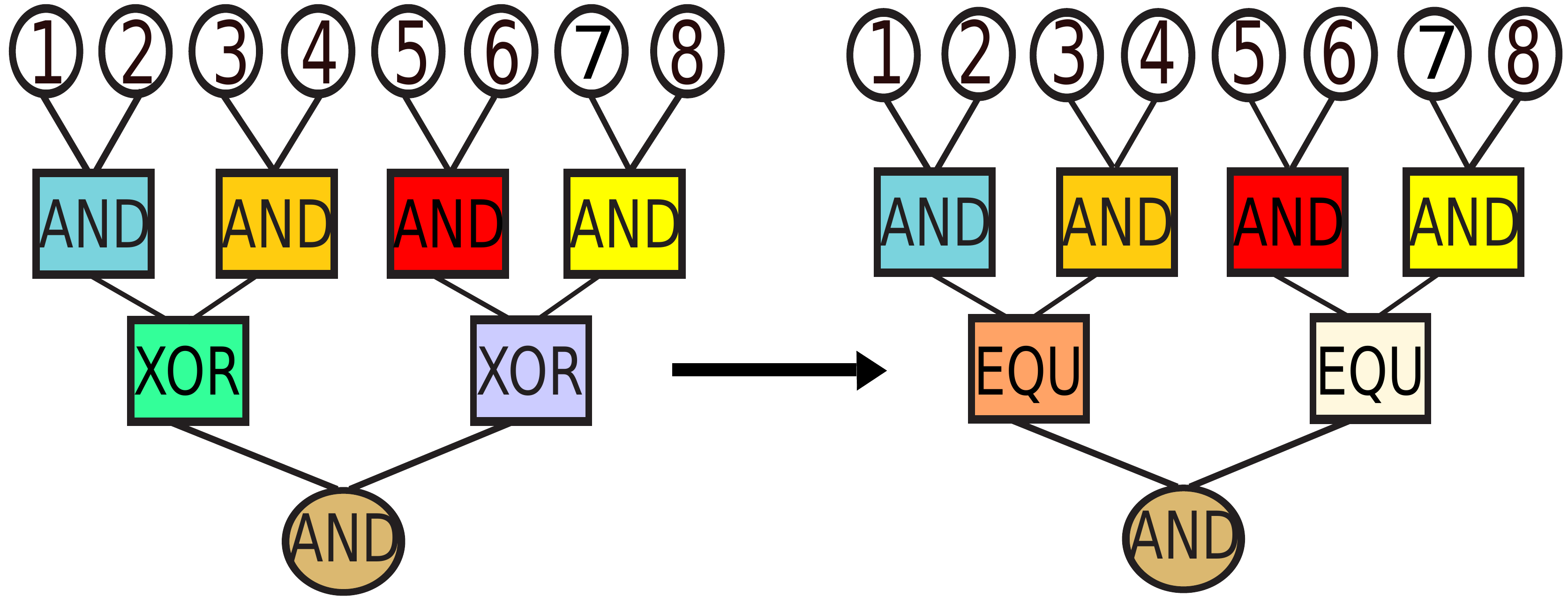}     
      \end{tabular}
    }
  \end{figure*}
    \begin{figure*}
    \centering
    \vspace{-10pt}
    \begin{minipage}[t]{0.49\textwidth}
       {\setlength{\tabcolsep}{0em}
          \begin{tabular}{ p{0.55\textwidth}}           
          \vspace{20pt}\hspace{-72pt}B. P\&CC-NonMod 
          \end{tabular}
          \begin{tabular}{ p{1.6\textwidth}}           
          \vspace{-10pt}\hspace{-70pt}\includegraphics[trim= 0.1cm 4cm 0cm 1.8cm,clip,scale=0.62]{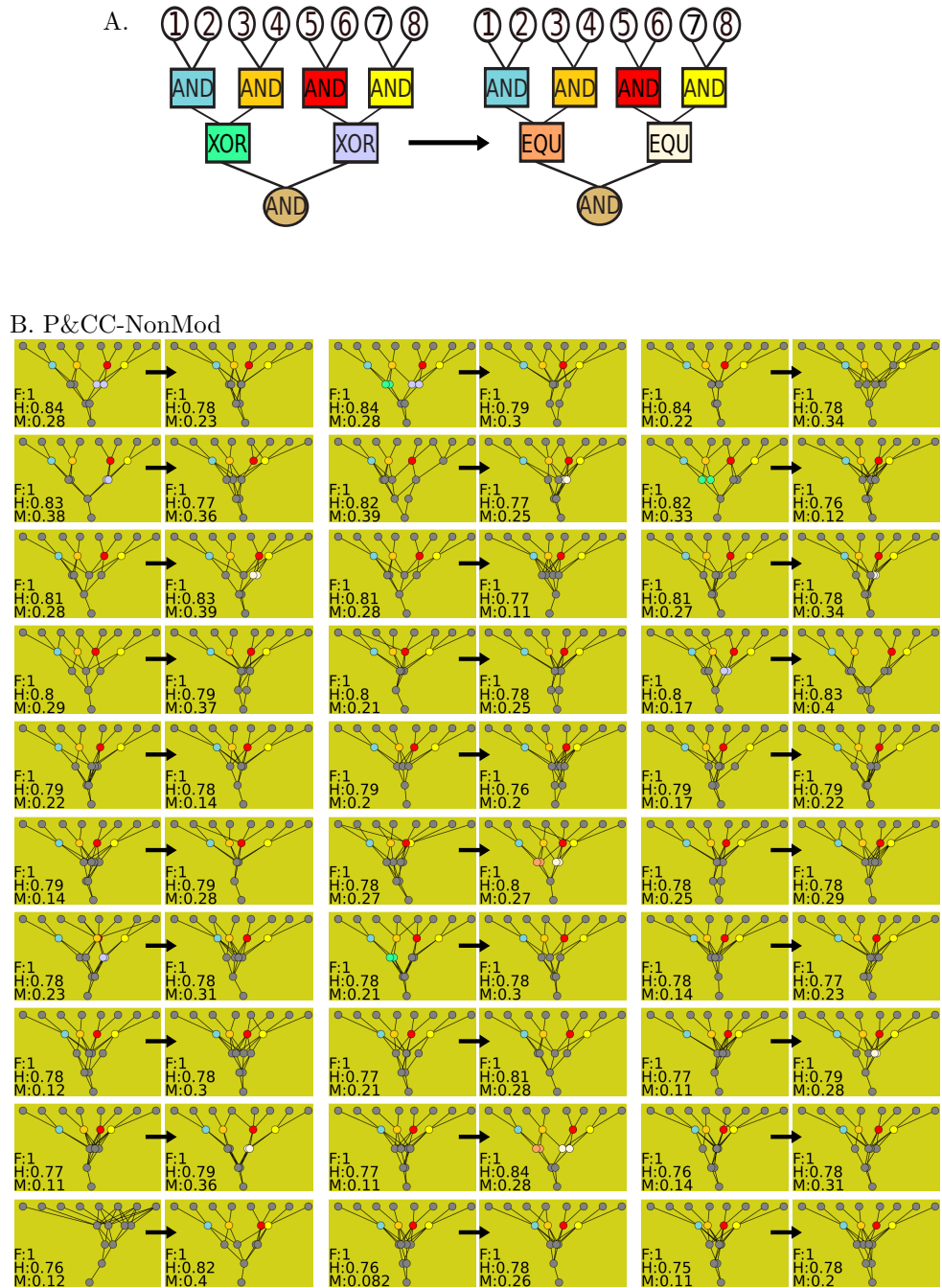}   
          \end{tabular}
        }
    \end{minipage}%
    \caption{Evolvability is improved even in networks that are hierarchical, but non-modular, demonstrating that the property of hierarchy conveys evolvability independent of modularity. \textbf{(A)} The base problem that networks originally evolved on (left) and the new, target problem that networks are transferred to and further evolved on (right). \textbf{(B)}     
In each pair, on the left is a perfect-performing network evolved for the base problem and on the right is an example descendant network that evolved on the target problem (specifically, the descendant network with median hierarchy). Except for being the P\&CC-NonMod treatment, this evolvability experiment has the same setup as the evolvability experiment in Fig. S\ref{fig:S5}.
    \label{fig:S11}}
    \end{figure*}
\pagebreak
%12
\begin{figure*}
\vspace{-50pt}
\centering
\includegraphics[scale = 0.5]{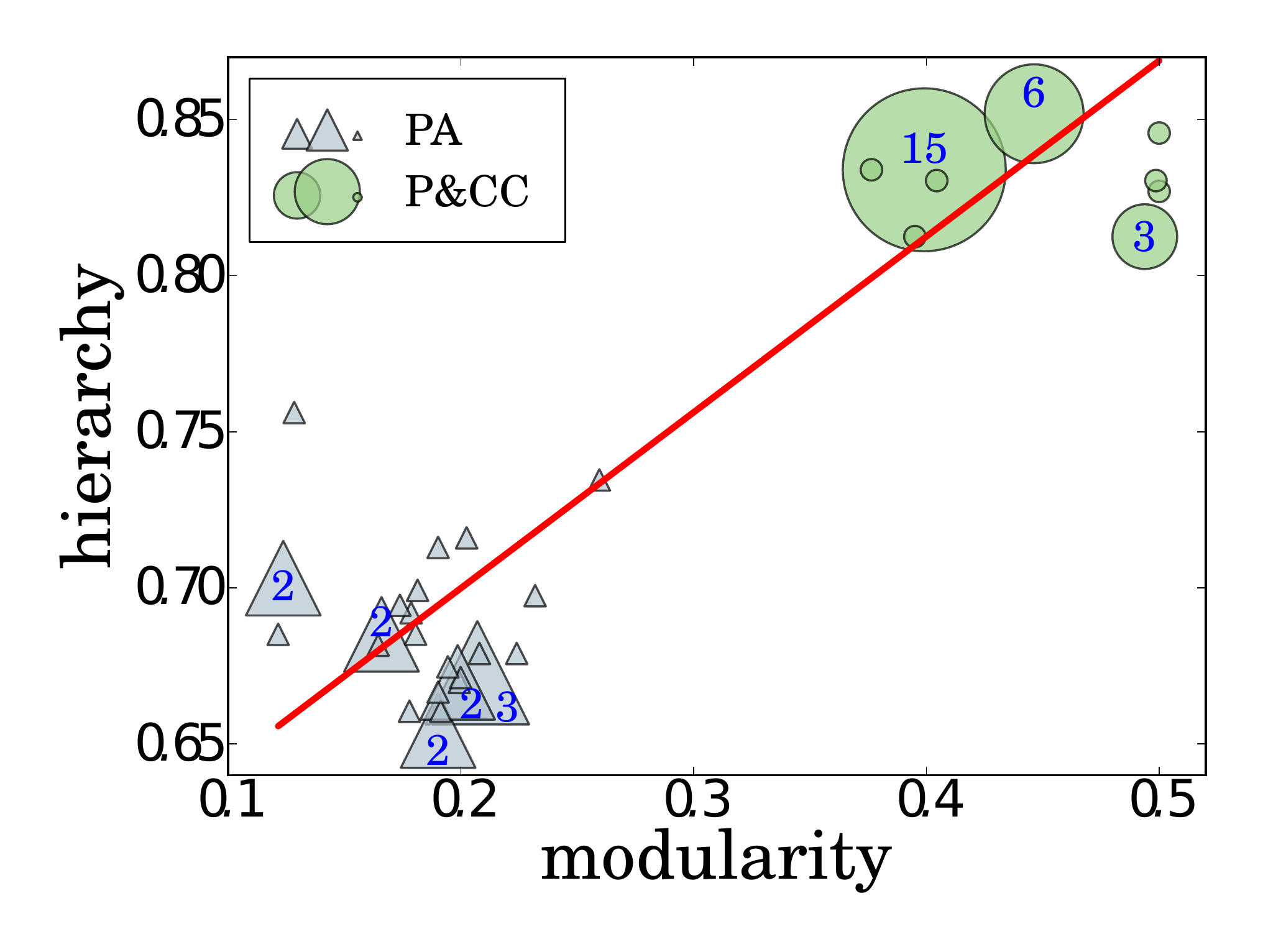}
\caption{There is a strong, linear, and positive correlation between network hierarchy and modularity. The Pearson's correlation coefficient is 0.92. The correlation is significant ($p < 0.00001$), as calculated by a t-test with a correlation of zero as the null hypothesis. Larger circles or triangles indicate the presence of more than one network at that location (the number describes how many).}
\label{fig:S12}
\end{figure*}

%13
\begin{figure*}
\centering
\subfigure{
\hspace{0pt}\includegraphics[scale=0.4]{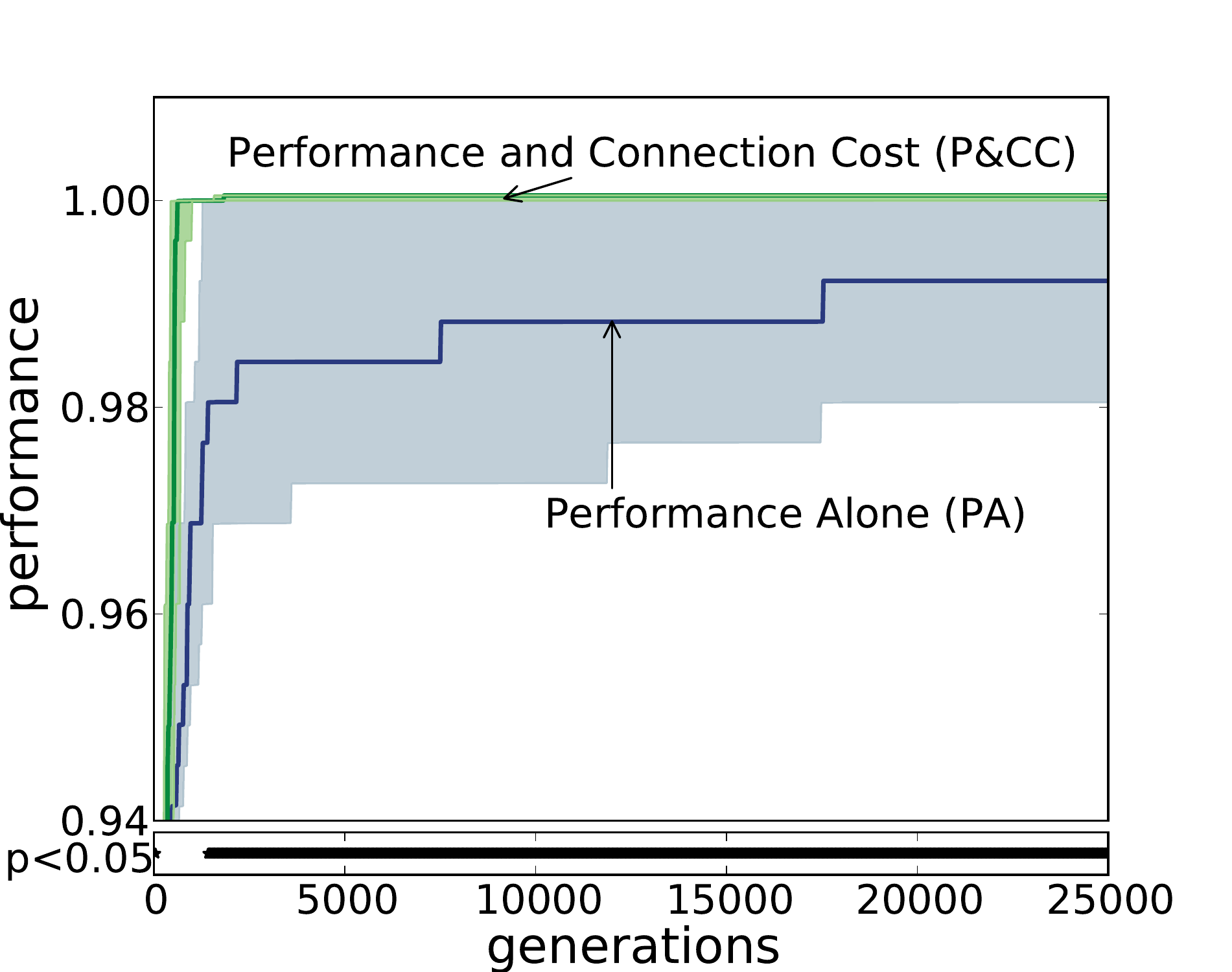}}
\subfigure{
\hspace{0pt}\includegraphics[scale=0.4]{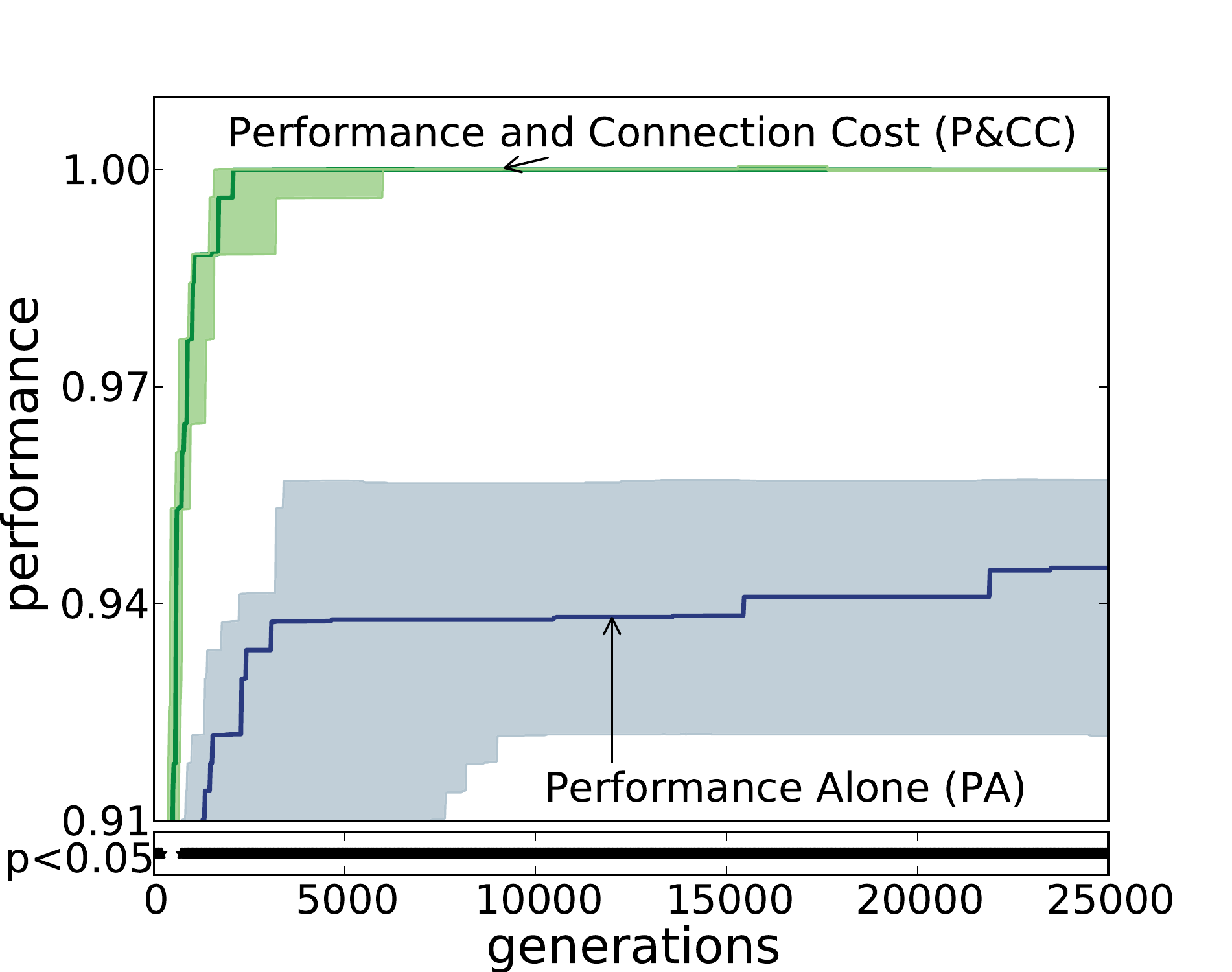}}
\caption{In addition to the main experimental problem, the P\&CC treatment also evolved high-performing networks faster than the PA treatment on two different problems (AND-EQU-AND, left, and OR-XOR/EQU-EQU, right; both are pictured in Fig.~8 in the main text). The bar below each plot indicates when a significant difference exists between the two treatments.}
\label{fig:PCC_reaches_faster_S13}%\vspace{-3.0em}
\end{figure*}

%14
\begin{figure}
\hspace{0pt}
\begin{minipage}[t]{0.5\textwidth}
{\setlength{\tabcolsep}{0em}
\begin{tabular}{p{0.097\textwidth} p{0.5\textwidth} p{0.097\textwidth} p{0.5\textwidth} p{0.097\textwidth}  p{0.4\textwidth}}          
\vspace{-80pt}\hspace{0pt}A&
\vspace{-100pt} \hspace{-10pt}\includegraphics[trim= 0.8cm 0.2cm 0.7cm 0.2cm,clip,scale=0.6]{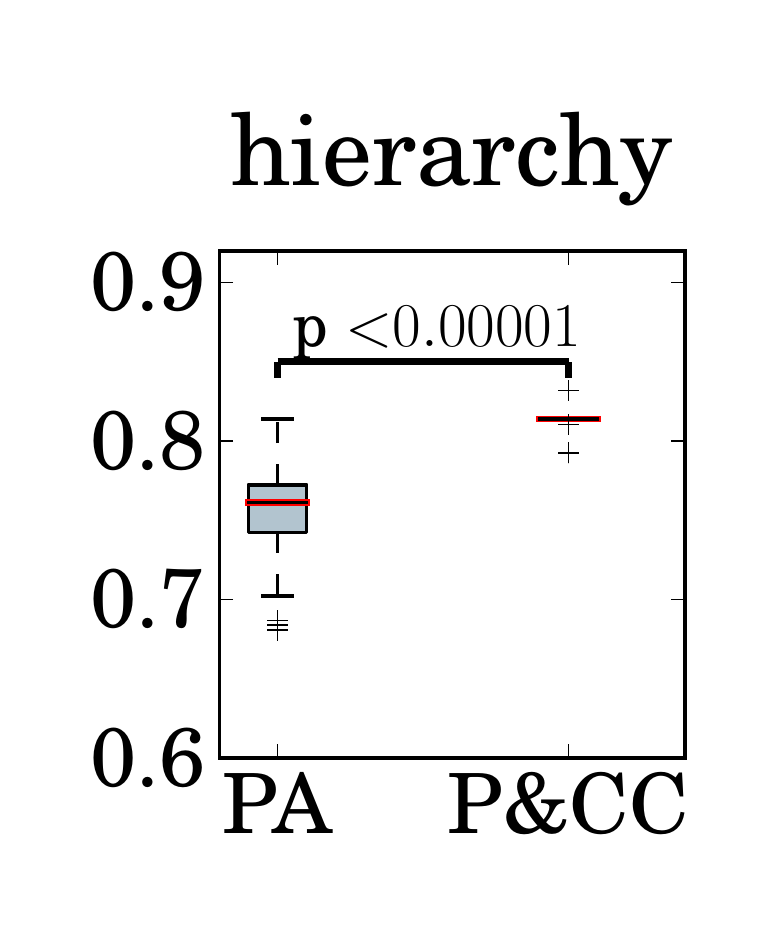}&  
     
\vspace{-80pt}B&
\vspace{-100pt} \includegraphics[trim= 0.7cm 0.2cm 0cm 0.2cm,clip,scale=0.6]{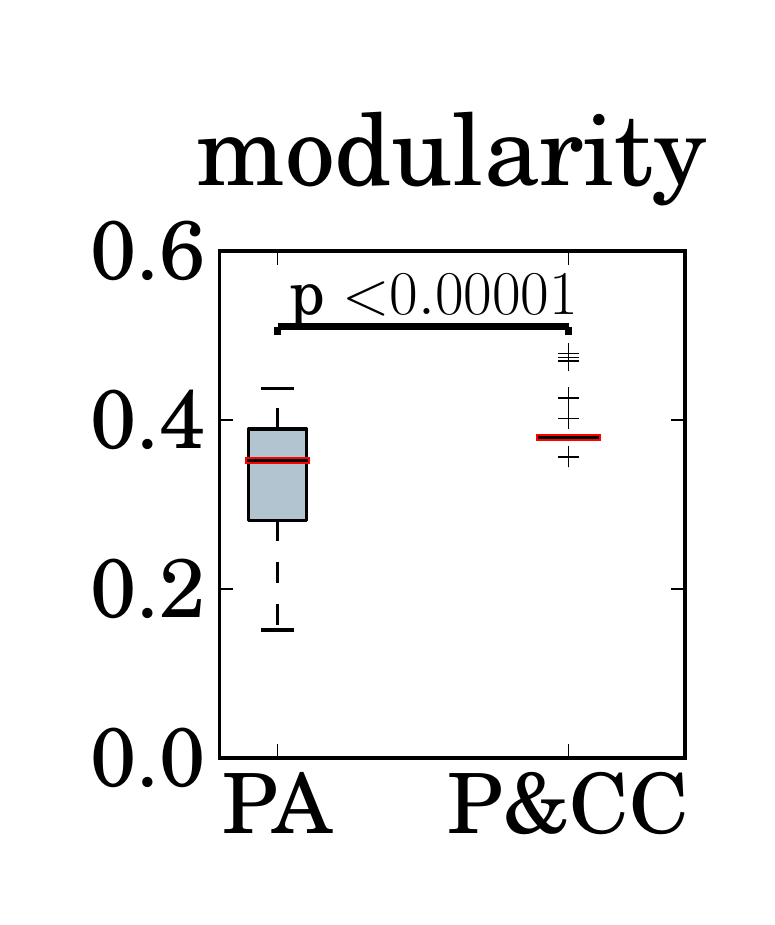}&
\vspace{-80pt}C&
\vspace{-100pt}\includegraphics[trim=  0cm 0.2cm 0.0cm 0.2cm,clip,scale=0.6]{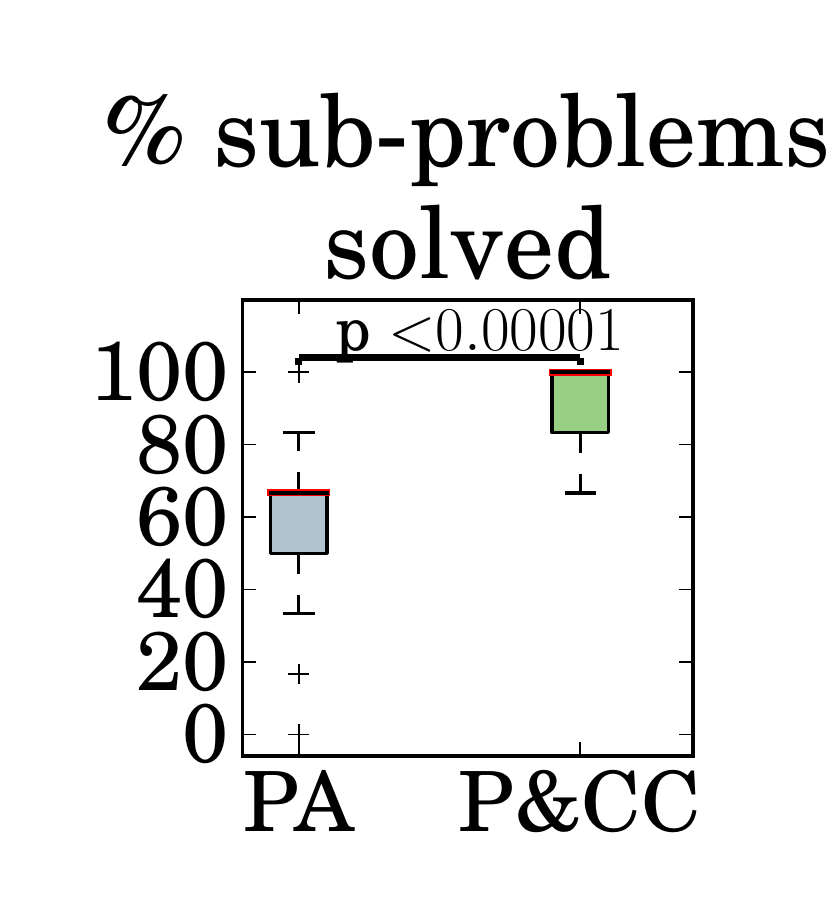}         
 \end{tabular}
}  
\end{minipage}
\caption{Our results are qualitatively unchanged when initializing networks with sparsely connected networks. In this experiment, the minimum and maximum number of initial connections that networks start with in generation 0 are 11 and 20, respectively. Due to the fact that at least 11 connections are needed to solve the experimental problem, networks that have an initial number of connections within this range are considered sparse (note: the default range for initial number of connections is [20, 100], Methods). The hierarchy \textbf{(A)}, modularity \textbf{(B)}, and percent of sub-problems solved \textbf{(C)} are significantly higher for end-of-run P\&CC networks, indicating that, regardless of the initial connectivity of networks, a connection cost promotes the evolution of these traits.}
\label{fig:S14}
\end{figure}
\begin{figure*}
%15
\begin{tabular}{p{0.007\textwidth} p{0.85\textwidth}}
\vspace{-140pt}A&
\includegraphics[width=0.9\textwidth]{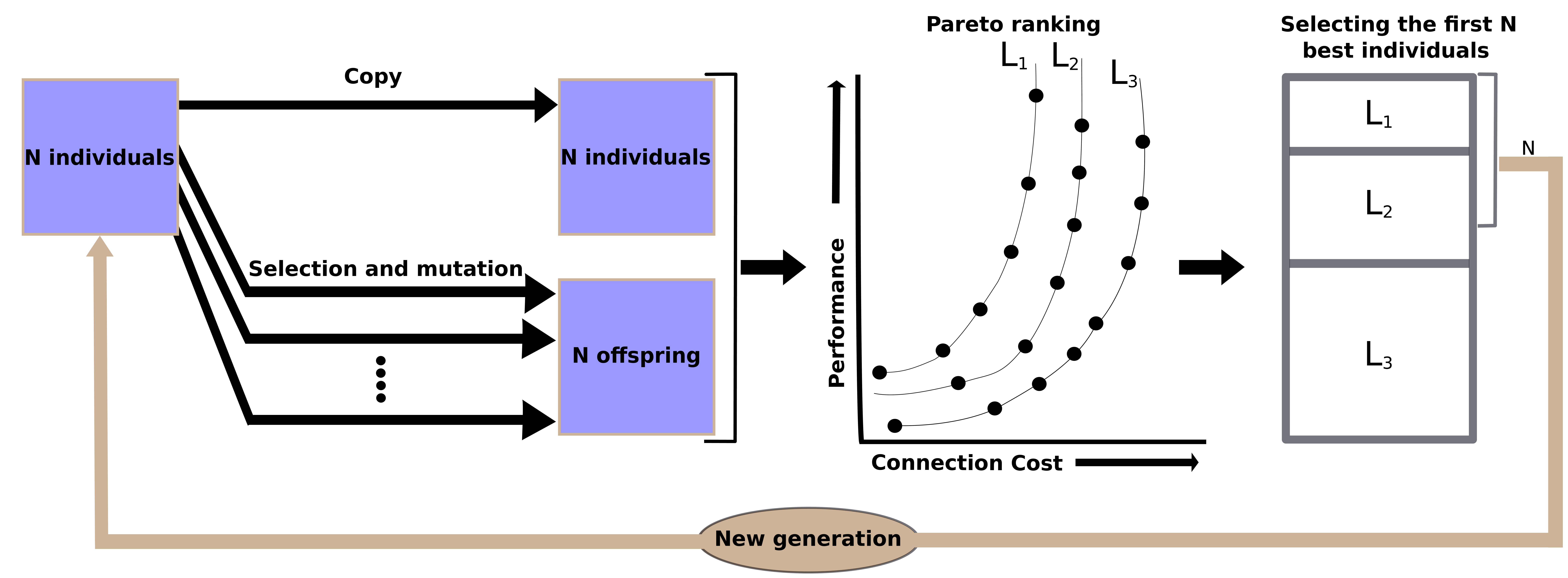} 
\end{tabular}
\hspace{-100pt}
\begin{tabular}{p{0.007\textwidth} p{0.45\textwidth} p{0.01\textwidth} p{0.4\textwidth}}
\vspace{-158pt}B&
\hspace{0pt}\includegraphics[width=0.4\textwidth]{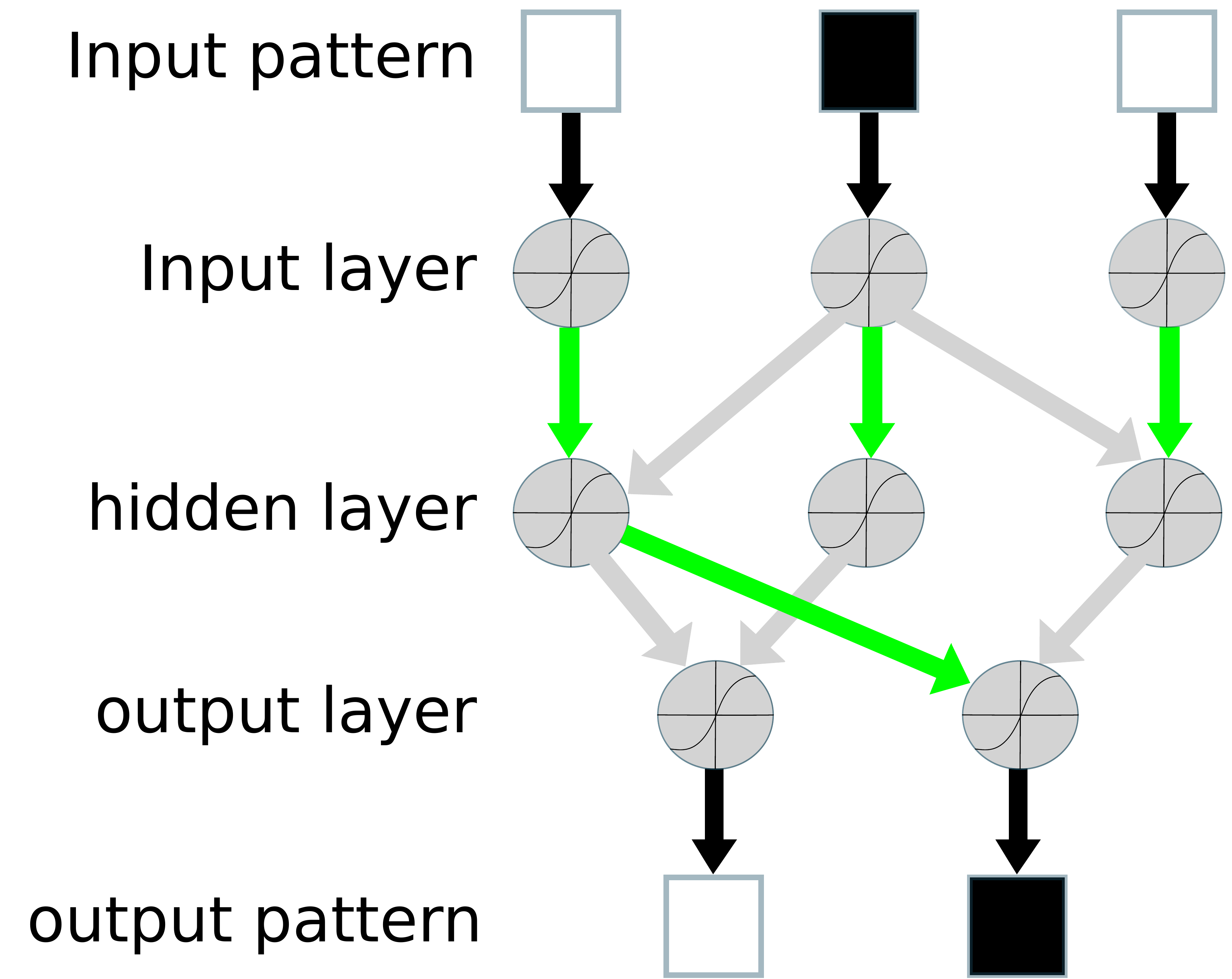}&
\vspace{-158pt}C&
\includegraphics[width=0.47\textwidth]{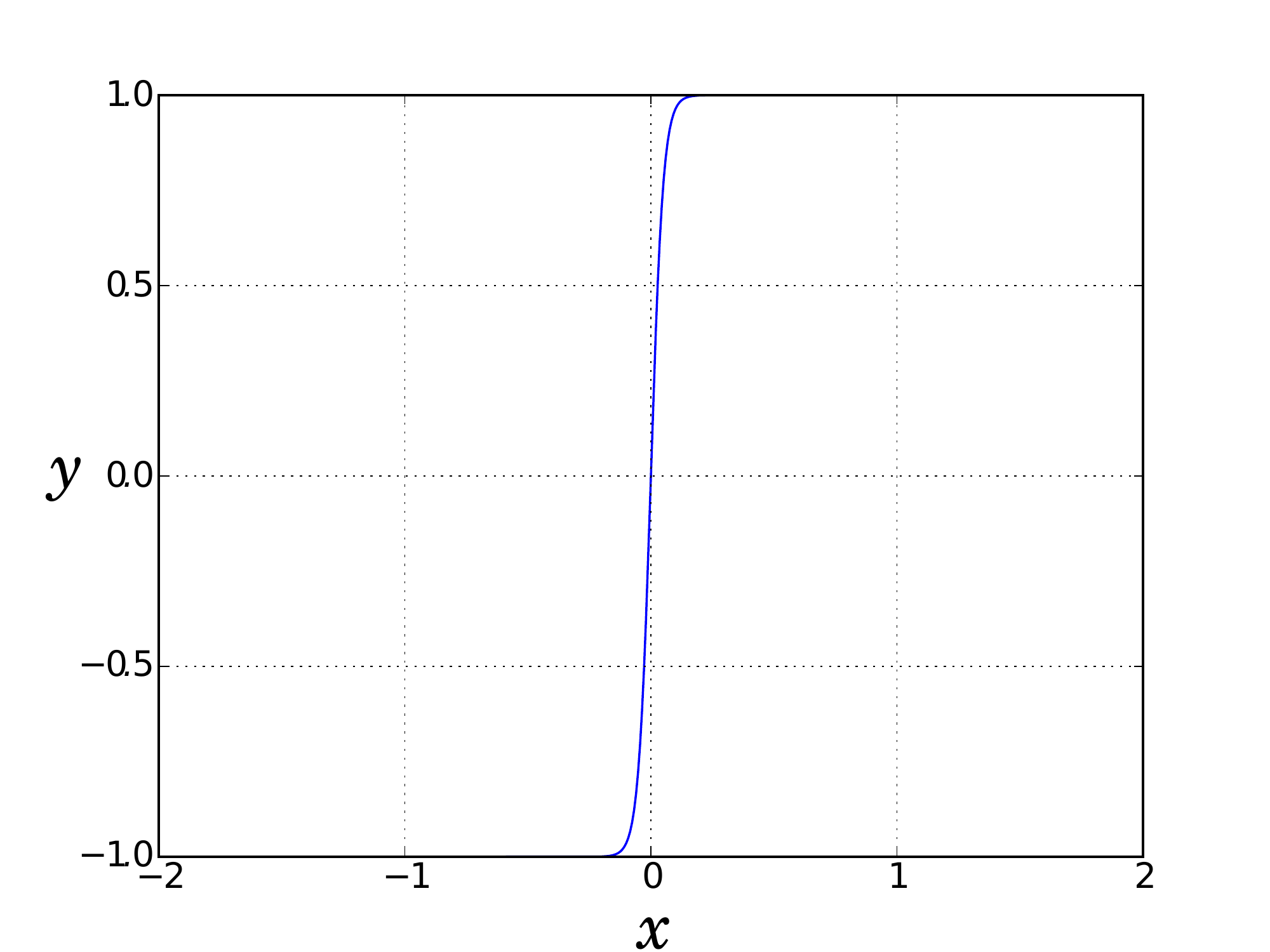}

\end{tabular}
\caption{Details of the evolutionary algorithm (figure adapted from \cite{clune2013:evolutionary}). \textbf{(A)} A graphical depiction of the the multi-objective evolutionary algorithm in our study, which is called the Non-dominated Sorting Genetic Algorithm version II (NSGA-II)\cite{Deb:2001}. In NSGA-II, evolution starts with a population of $N$ randomly generated networks. $N$ offspring are generated by randomly mutating the best of these individuals (as determined by tournament selection, wherein the best organism of 2 randomly selected organisms is chosen to produce offspring asexually). The combined pool of offspring and the current population are ranked based on Pareto dominance, and the best $N$ networks are selected to form the next generation. This process continues for a fixed number of generations or until networks with the desired performance or properties evolve. \textbf{(B)} An example network model. Networks are typically used by researchers to abstract the activities of many biological networks, such as gene regulatory networks and neural networks \cite{clune2013:evolutionary,Kashtan2005,Kashtan2007,geard2005gene,floreano:2008bio}. Nodes (analogous to neurons or genes) represent processing units that receive inputs from neighbors or external sources and process them to compute an output signal that is propagated to other nodes. For example, nodes at the input layer are activated by environmental stimuli and their output is passed to internal nodes. In this figure, arrows indicate a connection between two nodes, and thus illustrate the pathways through which information flows. Each connection has a weight, which is a number that controls the strength of interaction between the two nodes. Information flows through the network, ultimately determining the firing pattern of output nodes. The firing patterns of output nodes can be considered as commands that activate genes in a gene regulatory network or that move muscles in an animal body. The output value of each node, $y$, is a function of its weighted inputs and bias. In this paper, the specific \emph{activation function} is \textit{tanh(20x)}, where $x=\sum_i \left(w_i I_i + b\right)$, and where $I_i$ is the i\textsuperscript{th} input, $w_i$ the associated synaptic weight, and $b$ a bias that, like the weight vector, is evolved.  The specific function is depicted in \textbf{(C)}. Multiplying the input by 20 makes the function more like a step function. The output range is [-1,1].}
\label{fig:S15}
\end{figure*}

\end{document}